\documentclass[10pt,twocolumn,letterpaper]{article}

\usepackage[pagenumbers]{cvpr} 

\usepackage[dvipsnames]{xcolor}

\definecolor{ljh}{RGB}{180,120,120}

\definecolor{cvprblue}{rgb}{0.21,0.49,0.74}
\usepackage[pagebackref,breaklinks,colorlinks,citecolor=cvprblue]{hyperref}

\usepackage{graphicx}
\usepackage{amsmath}
\usepackage{amssymb}

\usepackage{tabu}

\usepackage{bm}
\usepackage{pifont}
\newcommand{\cmark}{\ding{51}}%
\usepackage{multirow}
\usepackage{colortbl}
\usepackage{arydshln}
\usepackage{makecell}
\usepackage{color}
\definecolor{Dred}{rgb}{0.6,0,0}
\definecolor{Dgreen}{rgb}{0,0.6,0}
\definecolor{Cup}{rgb}{0.6,0,0}
\definecolor{Cdown}{rgb}{0,0.6,0}
\definecolor{indomain}{RGB}{211, 230, 247}
\definecolor{id}{RGB}{211, 230, 247}
\definecolor{outdomain}{RGB}{255, 236, 236}
\definecolor{od}{RGB}{255, 236, 236}

\definecolor{Repeat}{rgb}{0.5,0.25,0}

\newcommand{\sm}{\textit{\textcolor{black}{supplementary materials}}}

\newtheorem{definition}{Definition}

\def\paperID{3120} 
\def\confName{CVPR}
\def\confYear{2024}

\title{Robust Synthetic-to-Real Transfer for Stereo Matching}

\author{
    Jiawei Zhang$^1$, Jiahe Li$^1$, Lei Huang$^2$, Xiaohan Yu$^3$, Lin Gu$^{4,5}$, Jin Zheng$^1$, Xiao Bai$^1$\thanks{Corresponding author: Xiao Bai (baixiao@buaa.edu.cn).}  \\
    $^1$School of Computer Science and Engineering, State Key Laboratory of \\ Complex \& Critical Software Environment,\, Jiangxi Research Institute,\, Beihang University\\
    $^2$SKLCCSE, Institute of Artificial Intelligence,  Beihang University \\
    $^3$School of Computing, Macquarie University, Australia
    \quad$^4$RIKEN AIP \quad$^5$The University of Tokyo
} 

\begin{document}
\maketitle

\begin{abstract}
With advancements in domain generalized stereo matching networks, models pre-trained on synthetic data demonstrate strong robustness to unseen domains. However, few studies have investigated the robustness after fine-tuning them in real-world scenarios, during which the domain generalization ability can be seriously degraded. In this paper, we explore fine-tuning stereo matching networks without compromising their robustness to unseen domains. Our motivation stems from comparing Ground Truth (GT) versus Pseudo Label (PL) for fine-tuning: GT degrades, but PL preserves the domain generalization ability. Empirically, we find the difference between GT and PL implies valuable information that can regularize networks during fine-tuning. We also propose a framework to utilize this difference for fine-tuning, consisting of a frozen Teacher, an exponential moving average (EMA) Teacher, and a Student network. The core idea is to utilize the EMA Teacher to measure what the Student has learned and dynamically improve GT and PL for fine-tuning. We integrate our framework with state-of-the-art networks and evaluate its effectiveness on several real-world datasets. Extensive experiments show that our method effectively preserves the domain generalization ability during fine-tuning. Code is available at: \url{https://github.com/jiaw-z/DKT-Stereo}.
\end{abstract}

\section{Introduction}
\label{sec: intro}

Estimating 3D geometry from 2D images is a fundamental problem in computer vision. Stereo matching is a solution that identifies matching correspondences and recovers depth information through triangulation. With the development of deep learning, stereo matching networks have shown impressive performance on various benchmarks.

\begin{figure}[t]
\centering
\includegraphics[width=1.0\linewidth]{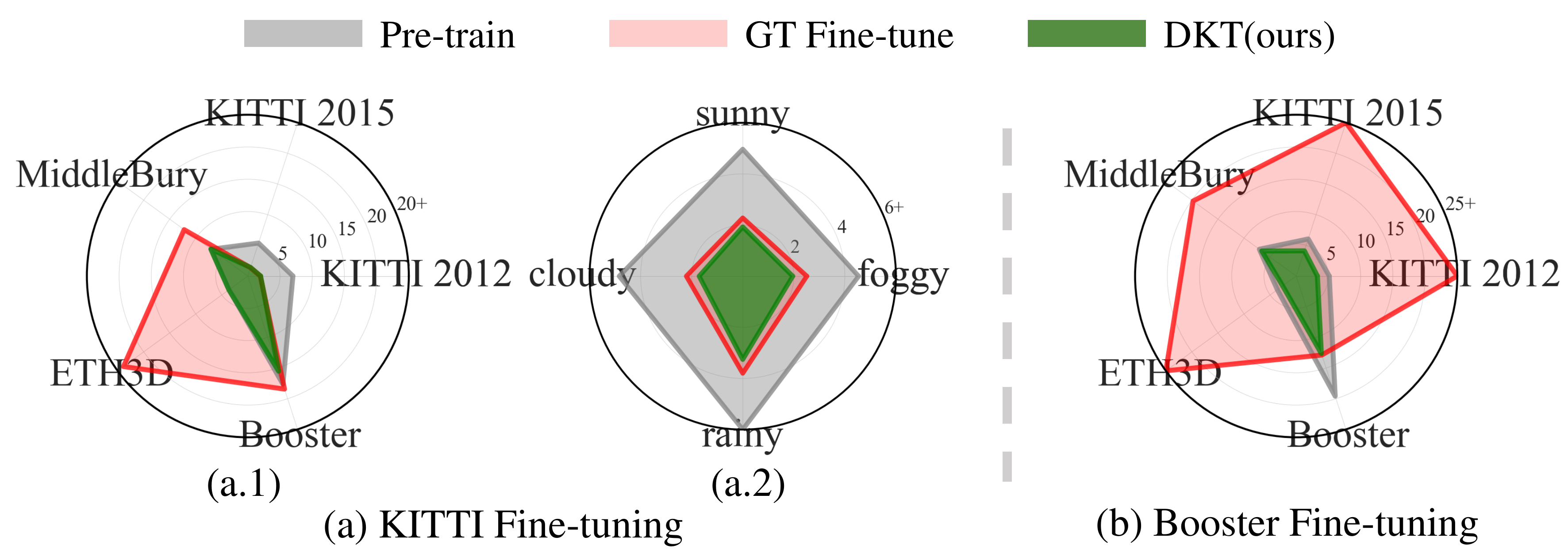}
\caption{Target-domain and cross-domain performance of networks pre-trained on synthetic data, fine-tuned with GT, and with our method (best visualized in colors). IGEV-Stereo \cite{xu2023iterative} is used as the baseline model. D1 error (the lower the better) is used for evaluation. (a) and (b) fine-tune networks on the KITTI and Booster datasets, respectively. Our method achieves great performance in target and unseen domains simultaneously. We also evaluate the robustness of KITTI fine-tuned networks on DrivingStereo in (a.2), where our method is more robust to challenging weather.}
\label{fig: intro}
\end{figure}

\begin{figure*}[t]
\setlength{\abovecaptionskip}{4pt}
\centering
\includegraphics[width=0.95\linewidth]{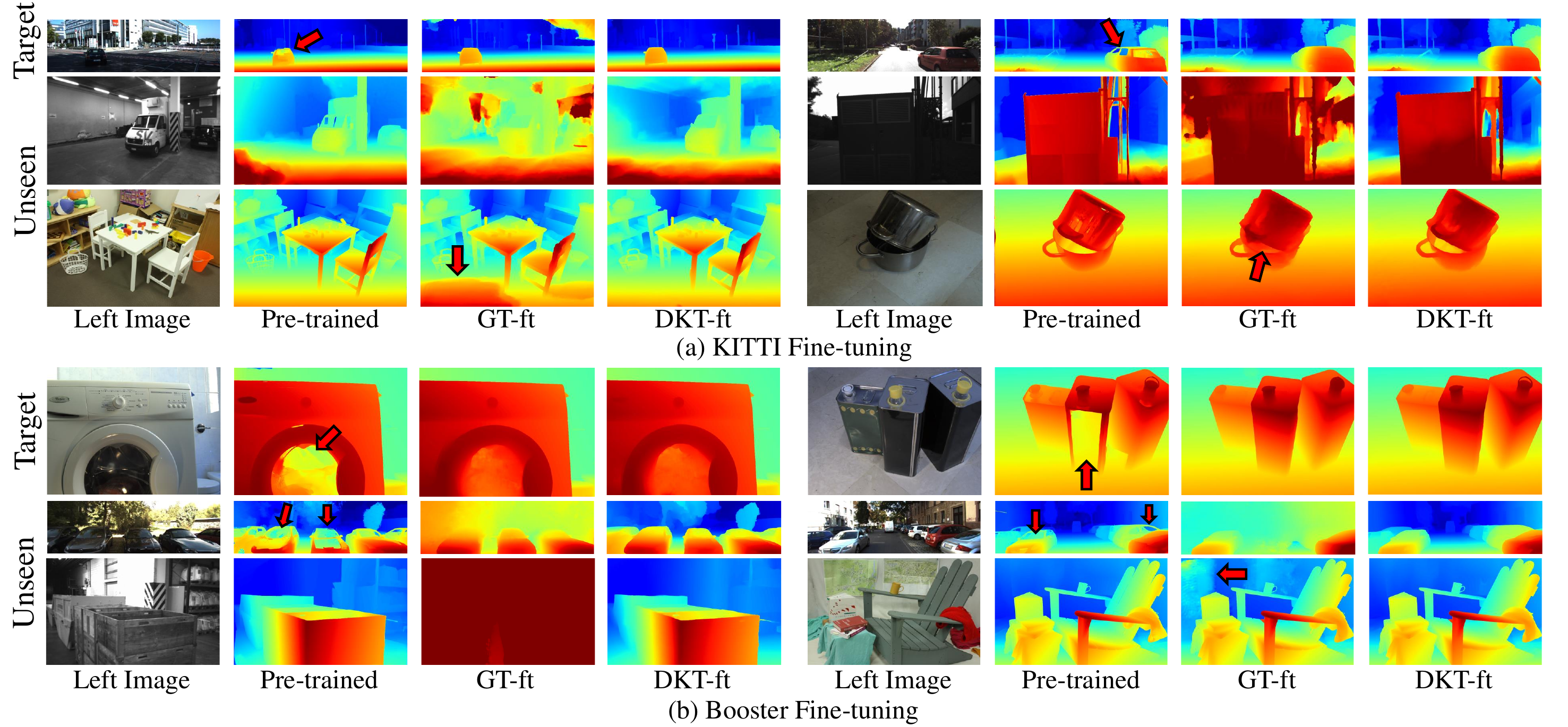}
\caption{Visualization results on both target and unseen domains. IGEV-Stereo \cite{xu2023iterative} is used as the baseline model. The network pre-trained on synthetic data shows robustness to unseen domains but can still fail to handle unseen challenges such as transparent or mirror (ToM) surfaces. Fine-tuning the network with GT improves the target-domain performance, however, it seriously degrades the domain generalization ability. Our DKT fine-tuning framework performs well on target and unseen domains simultaneously.}
\label{fig:intro_visualization}
\vspace{-2mm}
\end{figure*}

A major obstacle to their applications is the high cost of collecting real-world annotations. To make up for the lack of real data, existing networks are commonly pre-trained on synthetic data. With recent advancements in building domain generalized networks \cite{li2018domain,chuah2022itsa,zhang2022revisiting,lipson2021raft}, they have demonstrated strong robustness to unseen domains. However, as shown in \Cref{fig: intro}, fine-tuning pre-trained networks in real-world scenarios degrades the domain generalization ability. We provide visualization results in \Cref{fig:intro_visualization}. The degradation of robustness to unseen scenarios can render networks unreliable for real-world applications. To solve it, we explore what degrades the domain generalization ability of stereo networks during fine-tuning and provide a solution by combining Ground Truth (GT) with Pseudo Label (PL) predicted by pre-trained stereo matching networks.

Our exploration stems from the observation that using PL for fine-tuning stereo matching networks can preserve the domain generalization ability (\cf \Cref{sec: pseudo label preserves generalization ability}). Since there is always a gap between predicted PL and GT, we consider whether the difference between them implies certain knowledge. We are also motivated by previous studies \cite{hinton2015distilling,furlanello2018born,zhao2022decoupled} of knowledge distillation utilizing wrong response to non-target classes, which they refer to as ``dark knowledge'' of a transfer set by introducing PL with GT.

Starting from the observation, we aim to identify how GT and the predicted PL behave differently and further answer why fine-tuning with GT degrades the domain generalization ability of stereo matching networks. Specifically, we divide pixels into consistent and inconsistent regions based on the difference between GT and PL (\cf \Cref{sec: preliminary}). GT and PL of the consistent region are similar, otherwise, it is defined as the inconsistent region. We empirically find two primary causes that degrade the domain generalization ability: (\romannumeral1) In the inconsistent region, networks encounter new knowledge not learned during pre-training on synthetic data. Learning new knowledge without sufficient consistent region for regularization seriously degrades the domain generalization ability. (\romannumeral2) In the consistent region, stereo matching networks can still overfit the details of GT.

Based on the exploration, we propose the framework utilizing Dark Knowledge to Transfer (DKT) stereo matching networks. DKT consists of a frozen Teacher, an exponential moving average (EMA) Teacher, and a Student network, all initialized with the same pre-trained weights. We use the frozen Teacher to predict PL. The EMA Teacher is updated by the Student's weights, serving as a dynamic measure of consistent and inconsistent regions. With the EMA Teacher's prediction, we propose the Filter and Ensemble (F\&E) module to improve disparity maps. For GT, F\&E filters out the inconsistent region with a probability, resulting in the remaining GT that has a reduced inconsistent region, avoiding insufficient consistent region for regularization. It also performs an ensemble between GT and the EMA Teacher's prediction in the consistent region, adding fine-grained permutations to prevent networks from overfitting GT details. For PL, F\&E filters out the inconsistent region between PL and the EMA Teacher's prediction and ensembles them in the consistent region, enhancing the accuracy of PL predicted by the frozen Teacher. We train the Student with improved GT and PL jointly. After fine-tuning, we keep the Student for inference.

Our main contributions are as follows:
\begin{itemize}
    \item To the best of our knowledge, we make the first attempt to address the domain generalization ability degradation for fine-tuning stereo matching networks. We divide pixels into consistent and inconsistent regions based on the difference between ground truth and pseudo label and demonstrate their varied roles during fine-tuning. Our further analysis of their roles identifies two primary causes of the degradation: learning new knowledge without sufficient regularization and overfitting ground truth details.
    \item We propose the F\&E module to address these two causes, filtering out the inconsistent region to avoid insufficient regularization and ensembling disparities in the consistent region to prevent overfitting ground truth details.
    \item We introduce a dynamic adjustment for different regions by incorporating the exponential moving average Teacher, achieving the balance of preserving domain generalization ability and learning target domain knowledge.
    \item We develop the DKT fine-tuning framework, which can be easily applied to existing networks, significantly improving their robustness to unseen domains and achieving competitive target-domain performance simultaneously.
\end{itemize}
We believe this exploration will stimulate further consideration of stereo matching networks' robustness and domain generalization ability when fine-tuning them in real-world scenarios, crucial for their practical applications.

\section{Related Work}
\label{sec: related work}

\textbf{Stereo Matching Networks.} MC-CNN \cite{zbontar2015computing} first introduced a convolutional neural network to compute matching cost and predict disparity maps using SGM \cite{hirschmuller2005accurate}. Since then, numerous deep-learning-based methods have been developed for stereo matching \cite{kendall2017end,chang2018pyramid,cheng2020hierarchical,lipson2021raft,weinzaepfel2022improved}. According to their strategy to perform cost construction and aggregation, they can be divided into two categories. One type builds 3D cost volume and aggregates the cost with 2D convolutions. DispNetC \cite{mayer2016large} introduces end-to-end regression and builds a correlation-based 3D cost volume. Many works \cite{liang2018learning,yin2019hierarchical,tonioni2019real,xu2020aanet,poggi2021continual} adopt the correlation-based strategy and achieve impressive performance. Another type concatenates features to construct 4D cost volume and perform aggregation with 3D convolutions \cite{kendall2017end,chang2018pyramid,cheng2019learning,zhang2020adaptive,zhang2019ga,xu2022attention}. Methods in this category can leverage more comprehensive information from the original images and achieve leading performance on various benchmarks. Most recently, stereo matching networks \cite{lipson2021raft,wang2021pvstereo,xu2023iterative,zhao2023high,zeng2023parameterized} based on iterative optimization \cite{teed2020raft} are proposed and show state-of-the-art accuracy and strong robustness.

\textbf{Robust Stereo Matching.} Recently, building stereo matching networks that are robust to scenario changes has received increased interest. Existing studies can be categorized into joint generalization and cross-domain generalization types. Joint generalization involves training networks jointly on multiple datasets to perform well with a shared set of parameters \cite{liang2019stereo,shen2021cfnet,shen2023digging}. Cross-domain generalization aims to improve generalization performance on unseen scenarios, with a current focus on synthetic-to-real generalization \cite{zhang2020domain,cai2020matching,zhang2022revisiting,chuah2022itsa,rao2023masked,tosi2023nerf,chang2023domain}. These methods enhance performance in unseen domains by avoiding networks overfitting synthetic data during pre-training. However, they have not investigated the domain generalization ability after fine-tuning. This paper is unique as it attempts to address the domain generalization degradation during fine-tuning.

\textbf{Dark Knowledge.} Knowledge Distillation (KD) is explored for transferring knowledge from one teacher model to another student model \cite{hinton2015distilling,gou2021knowledge,yang2023multi}. In the original form for classification, it constructs a transfer set by using the soft target distribution produced by teachers. Dark knowledge refers to the extra information beyond one-hot GT contained in the modified training set \cite{sadowski2015deep},  associated with the distribution of responses to non-target classes. Building on this concept, a series of studies \cite{furlanello2018born,yuan2020revisiting,zhao2022decoupled} explore the effects of incorrect predictions in KD. BAN \cite{furlanello2018born} decomposes the KD into a dark knowledge term and ground truth component and develops a born-again procedure. DKD \cite{zhao2022decoupled} reformulates KD into target class and non-target class distillation and decouples the two parts to transfer dark knowledge. We adopt this concept to decouple different regions of disparity maps, analyzing what degrades the domain generalization ability of stereo matching networks during fine-tuning.

\section{Fine-tuning Stereo Matching Networks with Ground Truth and Pseudo Label}


\subsection{Preliminary and Definition} \label{sec: preliminary}

\begin{figure}[t]
\setlength{\abovecaptionskip}{4pt}
\centering
\includegraphics[width=1.0\linewidth]{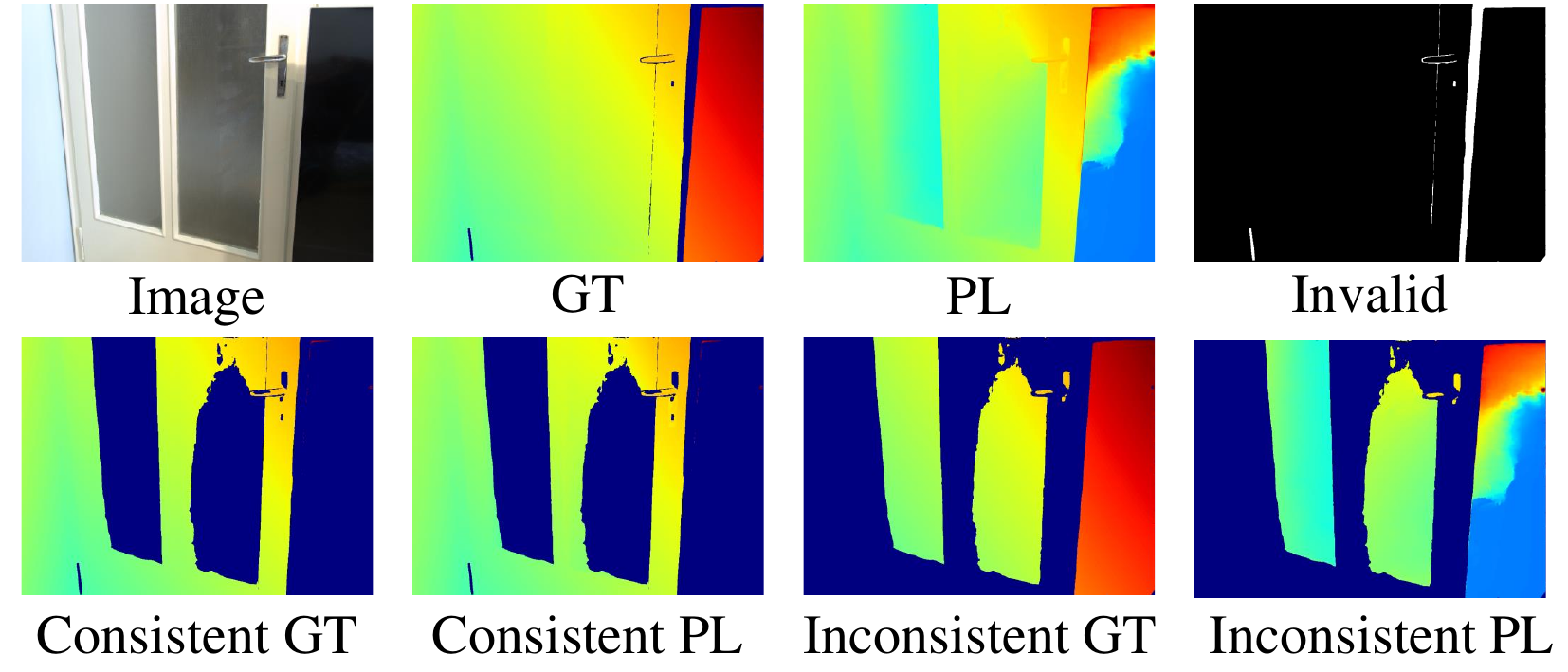}
\caption{Visualization of each region resulting from our division. We divide GT and PL based on their consistency.}
\label{fig: label decomposition}
\vspace{-.2cm}
\end{figure}

We explore whether stereo matching PL contains valuable knowledge that can preserve the domain generalization ability during fine-tuning and investigate the factors that make GT and PL behave differently. Stereo matching networks based on 2D or 3D cost aggregation are formed to obtain predictions by conducting Soft-Argmin on probability volume \cite{kendall2017end}, while the recently developed iterative optimization methods directly predict continuous values \cite{lipson2021raft,zhao2023high,xu2023iterative} and achieve superior robustness. The difference between GT and PL lies at the level of continuous values, making our study applicable to any form of stereo matching network. Given a rectified stereo pair $(I^{l}, I^{r})$ with GT map $D^{*}$, a pre-trained stereo matching network $\theta$ predicts a dense disparity map $\hat{D} = \theta(I^{l}, I^{r})$ as PL. We adopt the pixel distance $\hat{D} - D^{*}$ to measure responses to non-target disparities. To identify what makes GT and PL behave differently, we divide pixels into different regions based on the consistency between PL and GT. The visualization of each region is shown in \Cref{fig: label decomposition}. We define different regions as follows:
\begin{definition}
\vspace{-1mm}
\setlength{\abovedisplayskip}{0pt}
\setlength{\belowdisplayskip}{0pt}
    Consistent region $X_{c}(\tau)$. $X_{c}(\tau)$ contains pixels where the differences between GT and PL are less than the threshold $\tau$. $X_{c}(\tau) = \{x||\hat{D}(x_{i}) - D^{*}(x_{i})|<\tau\}$. This region represents areas where GT and PL align closely.
\end{definition}
\begin{definition}
\vspace{-1.5mm}
\setlength{\abovedisplayskip}{0pt}
\setlength{\belowdisplayskip}{0pt}
    Inconsistent region $X_{inc}(\tau)$. $X_{inc}(\tau)$ contains pixels where the differences between GT and PL are not less than the threshold $\tau$. $X_{inc}(\tau) = \{x||\hat{D}(x_{i}) - D^{*}(x_{i})|\geq\tau\}$. Stereo matching networks can encounter unseen challenges in $X_{inc}(\tau)$.
\end{definition}
\begin{definition}
\vspace{-1.5mm}
\setlength{\abovedisplayskip}{0pt}
\setlength{\belowdisplayskip}{0pt}
    Invalid region $X_{invalid}$. It contains pixels without available annotations due to the sparsity of GT.
\end{definition}

\subsection{Experiment Setup} \label{sec: experiment setup}

\textbf{Dataset.} Our experiments are conducted on KITTI \cite{geiger2012we,menze2015object}, DrivingStereo \cite{yang2019drivingstereo}, Booster \cite{ramirez2022open}, Middlebury \cite{scharstein2014high}, and ETH3D \cite{schops2017multi}. KITTI and DrivingStereo datasets collect outdoor driving scenes, Booster and Middlebury provide indoor scenes, and ETH3D includes indoor and outdoor scenes in grayscale. We use the half resolution of Middlebury and DrivingStereo, and the quarter resolution of Booster. Except for online submissions, we split datasets for local evaluation due to the submission policy. More details are shown in \sm.  In our tables, we use \textcolor{blue}{blue} for target domains and \textcolor{red}{red} for unseen domains.

\textbf{Metric.} We calculate the percentage of pixels with an absolute error larger than a certain distance for evaluation. 3 pixel is used for KITTI and DrivingStereo datasets. 2 pixel is used for Middlebury and Booster datasets. 1 pixel is used for ETH3D dataset.

\textbf{Implementation.} We adopt IGEV-Stereo \cite{xu2023iterative} as the basic network architecture here, and results of other networks are shown in \sm. We initialize networks with SceneFlow \cite{mayer2016large} pre-trained weights. For KITTI datasets, we fine-tune networks for 50k steps with a batch size of 8. For the Booster dataset, we fine-tune networks for 25k steps with a batch size of 2. We set the learning rate to 2e-4 for KITTI and 1e-5 for Booster. We use the data augmentation strategy in \cite{lipson2021raft} during fine-tuning.

\subsection{Ground Truth versus Pseudo Label} \label{sec: ground truth versus pseudo label}

\subsubsection{PL Preserves Domain Generalization Ability} \label{sec: pseudo label preserves generalization ability}

\begin{figure}[t]
\setlength{\abovecaptionskip}{2pt}
\centering
\includegraphics[width=1.0\linewidth]{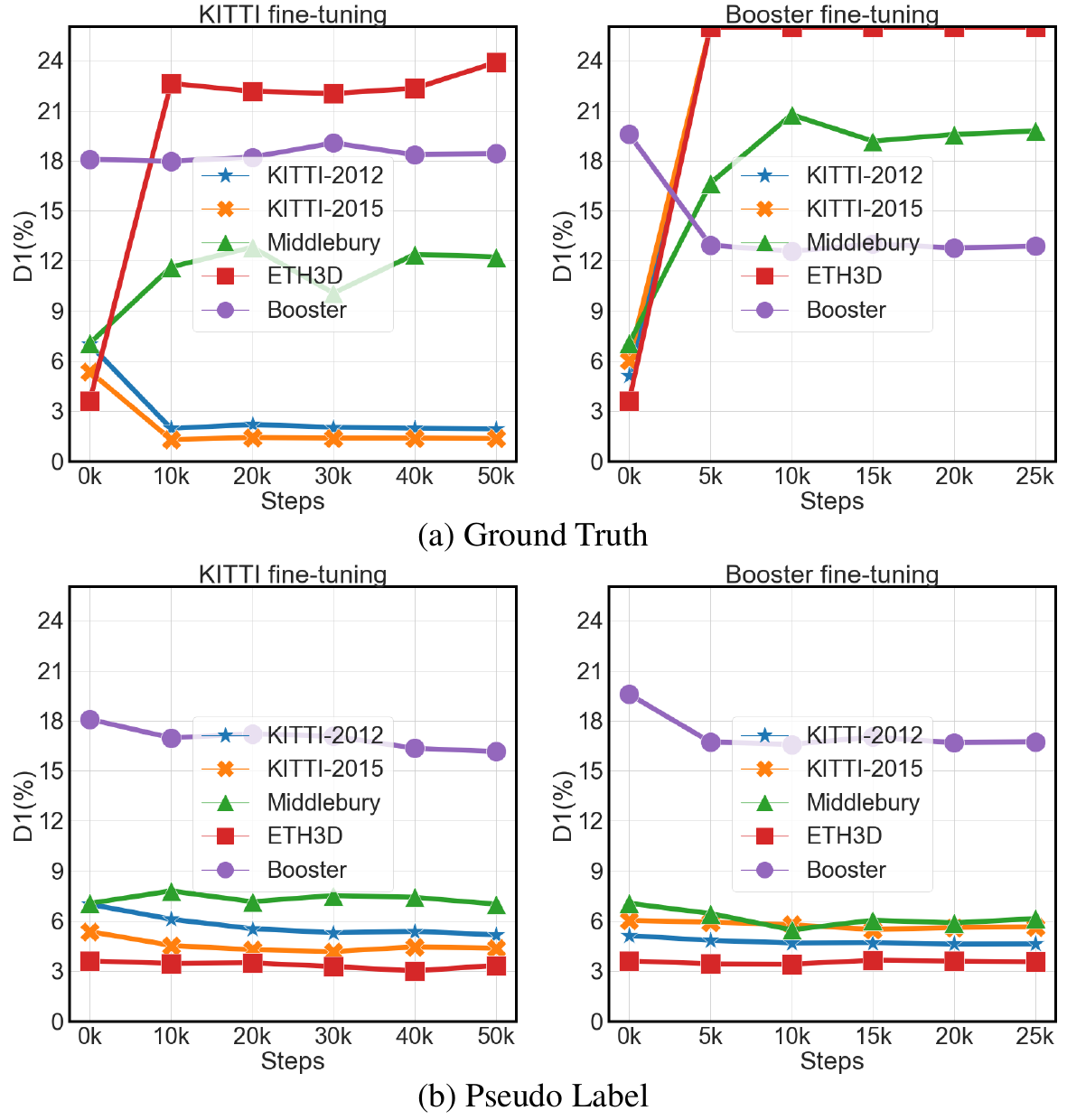}
\caption{Evaluation curves of the target and cross domain performance during fine-tuning with GT or PL.}
\vspace{-4mm}
\label{fig: PL preserve generalization ability}
\end{figure}

We compare the fine-tuning of pre-trained networks using GT and PL. Contrast observations lead us to believe that the difference between GT and PL contains valuable information. Analyzing the validation curves presented in \Cref{fig: PL preserve generalization ability}, we observe a significant degradation of domain generalization ability during GT fine-tuning, occurring at an early stage. In contrast, there is no substantial degradation in domain generalization ability when using PL. This finding demonstrates that PL, as naive knowledge produced by pre-trained stereo matching networks, can preserve the original domain generalization ability.

\vspace{-2mm}
\subsubsection{Investigating the Difference between GT and PL} \label{sec: approaching dark knowledge}

To understand the role of each region in GT and PL and to delve into the reasons behind the degradation of domain generalization ability, we conduct fine-tuning experiments by isolating different regions. The comparative results across various settings are presented in \Cref{table: explore dark knowledge}.

\textbf{Learning new knowledge without sufficient regularization can degrade domain generalization ability.} We start by comparing the baseline, which uses all valid regions of GT, with a setting that uses only the consistent region $X_{c}(3)$.  By removing the inconsistent region $X_{inc}(3)$ of GT, we observe an effective alleviation of generalization degradation, particularly notable for the Booster dataset. This observation highlights the significant impact of GT in the inconsistent region on degrading generalization ability. Next, we isolate only the inconsistent region $X_{inc}(3)$ for fine-tuning, where we observe a catastrophic degradation in generalization ability. Networks trained with only GT($X_{inc}(3)$) even suffer difficulties generalizing to the target-domain validation data. This comparison highlights the crucial regularization role played by GT in the consistent region for networks learning to address unseen challenges in the inconsistent region. We conclude that while GT supervision in the inconsistent region is valuable for networks adapting to unseen challenges, learning new knowledge in this region without sufficient regularization can seriously degrade domain generalization ability.

\textbf{Fine-grained details in GT can cause overfitting, degrading domain generalization ability.} Unlike the inconsistent region, where handling is more challenging, networks find handling the consistent region less difficult since most necessary knowledge has been involved in synthetic data. Subsequently, we investigate whether learning in $X_{c}(\tau)$ can also result in domain generalization degradation. In our observations, a clear degradation in domain generalization ability occurs when using GT compared to PL in the same consistent region $X_{c}(3)$. Further tightening the threshold for the consistent region to $X_{c}(1)$ provides some relief, however, we find that even inaccurate components within a one-pixel range can lead to varying domain generalization abilities. We attribute this to stereo matching networks overfitting fine-grained details in GT, which negatively impacts domain generalization ability.

\begin{table}[t]\footnotesize
\centering
\begin{tabular}{c|ccccc}
\hline
Supervision & 2012 & 2015 & Midd & ETH3D & Booster \\
\hline
\hline
Training set & \multicolumn{5}{c}{\textbf{KITTI 2012 \& 2015}} \\
\hline
zero-shot & 7.00 & 5.37 & 7.06 & 3.61 & 17.62 \\
\hdashline 
GT(valid) & \cellcolor{indomain} \textbf{1.94} & \cellcolor{indomain} \textbf{1.36} & \cellcolor{outdomain} 12.23 & \cellcolor{outdomain} 23.88 & \cellcolor{outdomain} 18.43 \\
GT($X_{c}(3)$) & \cellcolor{indomain} 2.17 & \cellcolor{indomain} 1.69 & \cellcolor{outdomain} 10.97 & \cellcolor{outdomain} 19.44 & \cellcolor{outdomain} 17.39 \\
GT($X_{inc}(3)$) & \cellcolor{indomain} 16.78 & \cellcolor{indomain} 22.01 & \cellcolor{outdomain} 28.06 & \cellcolor{outdomain} 69.88 & \cellcolor{outdomain} 36.50 \\
GT($X_{c}(1)$) & \cellcolor{indomain} 2.58 & \cellcolor{indomain} 2.07 & \cellcolor{outdomain} 9.89 & \cellcolor{outdomain} 11.76 & \cellcolor{outdomain} 17.95 \\
\hdashline
PL(all) & \cellcolor{indomain} 5.18 & \cellcolor{indomain} 4.37 & \cellcolor{outdomain} \textbf{7.01} & \cellcolor{outdomain} \textbf{3.34} & \cellcolor{outdomain} \textbf{16.15} \\
PL(valid) & \cellcolor{indomain} 5.82 & \cellcolor{indomain} 4.90 & \cellcolor{outdomain} 8.30 & \cellcolor{outdomain} 3.65 & \cellcolor{outdomain} 16.46 \\
PL($X_{c}(3)$) & \cellcolor{indomain} 3.72 & \cellcolor{indomain} 3.11 & \cellcolor{outdomain} 8.46 & \cellcolor{outdomain} 3.53 & \cellcolor{outdomain} 16.17 \\
PL($X_{inc}(3)$) & \cellcolor{indomain} 9.27 & \cellcolor{indomain} 8.64 & \cellcolor{outdomain} 11.80 & \cellcolor{outdomain} 11.67 & \cellcolor{outdomain} 28.25 \\
PL($X_{c}(1)$) & \cellcolor{indomain} 3.29 & \cellcolor{indomain} 2.92 & \cellcolor{outdomain} 8.10 & \cellcolor{outdomain} 3.69 & \cellcolor{outdomain} 16.26 \\
\hline
Training set & \multicolumn{5}{c}{\textbf{Booster}} \\
\hline
zero-shot & 5.13 & 6.04 & 7.06 & 3.61 & 19.49 \\
\hdashline 
GT(valid) & \cellcolor{outdomain} 52.30 & \cellcolor{outdomain} 55.44 & \cellcolor{outdomain} 19.78 & \cellcolor{outdomain} 93.31 & \cellcolor{indomain} \textbf{12.88} \\
GT($X_{c}(3)$) & \cellcolor{outdomain} 4.77 & \cellcolor{outdomain} 7.51 & \cellcolor{outdomain} 7.26 & \cellcolor{outdomain} 18.39 & \cellcolor{indomain} 14.07 \\
GT($X_{inc}(3)$) & \cellcolor{outdomain} 93.35 & \cellcolor{outdomain} 96.38 & \cellcolor{outdomain} 50.09 & \cellcolor{outdomain} 99.24 & \cellcolor{indomain} 30.25 \\
GT($X_{c}(1)$) & \cellcolor{outdomain} 4.58 & \cellcolor{outdomain} 7.66 & \cellcolor{outdomain} 6.93 & \cellcolor{outdomain} 17.15 & \cellcolor{indomain} 13.78 \\
\hdashline
PL(all) & \cellcolor{outdomain} 4.64 & \cellcolor{outdomain} 5.65 & \cellcolor{outdomain} 6.14 & \cellcolor{outdomain} 3.56 & \cellcolor{indomain} 16.74 \\
PL(valid) & \cellcolor{outdomain} 4.61 & \cellcolor{outdomain} 5.79 & \cellcolor{outdomain} 6.61 & \cellcolor{outdomain} 3.47 & \cellcolor{indomain} 16.87 \\
PL($X_{c}(3)$) & \cellcolor{outdomain} \textbf{3.23} & \cellcolor{outdomain} \textbf{4.36} & \cellcolor{outdomain} 6.15 & \cellcolor{outdomain} 3.28 & \cellcolor{indomain} 14.79 \\
PL($X_{inc}(3)$) & \cellcolor{outdomain} 5.70 & \cellcolor{outdomain} 6.53 & \cellcolor{outdomain} 7.35 & \cellcolor{outdomain} 4.55 & \cellcolor{indomain} 21.75 \\
PL($X_{c}(1)$) & \cellcolor{outdomain} 3.63 & \cellcolor{outdomain} 4.89 & \cellcolor{outdomain} \textbf{6.00} & \cellcolor{outdomain} \textbf{3.01} & \cellcolor{indomain} 14.26 \\
\hline
\end{tabular}
\caption{Results of fine-tuning stereo matching networks using different regions of GT or PL.}
\label{table: explore dark knowledge} 
\end{table}

\textbf{PL in the consistent region contains valuable knowledge to preserve the domain generalization ability.} We compare using all PL with only using the consistent region $X_{c}(3)$ of PL. Fine-tuning with PL($X_{c}(3)$) preserves the domain generalization ability. It indicates that details in PL do not contribute to domain generalization degradation, which we consider a positive distinction from the degradation of using GT supervision. Additionally, we observe enhanced performance in the target domain, attributed to the increased accuracy of PL($X_{c}(3)$) compared to PL(all).

\textbf{PL in the inconsistent region negatively impacts target-domain and unseen-domain performances.} In the inconsistent region, PL is considered incorrect relative to GT. We explore whether incorrect predictions follow any useful patterns. Compared to using the correct PL($X_{c}(3)$), training networks with PL($X_{inc}(3)$) degrades both target and unseen domain performances. Consequently, we conclude that PL in the inconsistent region has negative impacts, and keeping the correct PL is still vital.

\textbf{PL in the invalid region benefits domain generalization ability.} The invalid region $X_{invalid}$ contains the potential consistent region and inconsistent region for PL that we cannot directly distinguish due to the unavailable of GT. Based on the conclusion that the inconsistent region contributes negatively, we question whether invalid regions are still necessary and conduct comparisons between using PL(all) and PL(valid). For KITTI datasets, we observe additional benefits when introducing invalid regions, demonstrating utilizing PL in the valid region can be helpful. For the Booster dataset, we note that GT annotations are very dense and have fewer invalid regions, thus the effect is not that obvious.

\subsubsection{PL as Regularization with GT} \label{sec: dark knowledge as regularization with GT}

This section explores the role of PL as a regularization term when combined with GT for fine-tuning. We use PL of different regions with GT and make comparisons between different settings. The results are presented in \Cref{table: dark knowledge as regularization}.

\textbf{Using all regions of PL as a naive strategy.} We first fine-tune networks jointly with all available regions of GT and PL. While this approach mitigates the degradation in domain generalization, it exhibits inferior performance on target domains compared to using only GT, and its generalization is poorer than using only PL in \Cref{table: explore dark knowledge}. It indicates that this straightforward combination fails to leverage the advantages of both GT and PL.

\begin{table}[t]\footnotesize
\centering
\begin{tabular}{c|ccccc}
\hline
Supervision & 2012 & 2015 & Midd & ETH3D & Booster \\
\hline
\hline
Training set & \multicolumn{5}{c}{\textbf{KITTI 2012 \& 2015}} \\
\hline
GT & \cellcolor{indomain} \textbf{1.94} & \cellcolor{indomain} \textbf{1.36} & \cellcolor{outdomain} 12.23 & \cellcolor{outdomain} 23.88 & \cellcolor{outdomain} 18.43 \\
\hdashline 
GT+PL(all) & \cellcolor{indomain} 3.24 & \cellcolor{indomain} 2.75 & \cellcolor{outdomain} \textbf{7.40} & \cellcolor{outdomain} \textbf{3.64} & \cellcolor{outdomain} \textbf{16.54} \\
GT+PL($X_{c}(3)$) & \cellcolor{indomain} 2.14 & \cellcolor{indomain} 1.44 & \cellcolor{outdomain} 8.82 & \cellcolor{outdomain} 8.14 & \cellcolor{outdomain} 17.09 \\
GT+PL($X_{c}(1)$) & \cellcolor{indomain} 1.97 & \cellcolor{indomain} 1.38 & \cellcolor{outdomain} 8.47 & \cellcolor{outdomain} 11.29 & \cellcolor{outdomain} 17.75 \\
\hline
Training set & \multicolumn{5}{c}{\textbf{Booster}} \\
\hline
GT & \cellcolor{outdomain} 52.30 & \cellcolor{outdomain} 55.44 & \cellcolor{outdomain} 19.78 & \cellcolor{outdomain} 93.31 & \cellcolor{indomain} 12.88 \\
\hdashline 
GT+PL(all) & \cellcolor{outdomain} \textbf{7.14} & \cellcolor{outdomain} \textbf{10.69} & \cellcolor{outdomain} \textbf{8.02} & \cellcolor{outdomain} \textbf{10.30} & \cellcolor{indomain} 14.29 \\
GT+PL($X_{c}(3)$) & \cellcolor{outdomain} 41.95 & \cellcolor{outdomain} 45.27 & \cellcolor{outdomain} 11.74 & \cellcolor{outdomain} 65.63 & \cellcolor{indomain} \textbf{12.18} \\
GT+PL($X_{c}(1)$) & \cellcolor{outdomain} 45.77 & \cellcolor{outdomain} 48.16 & \cellcolor{outdomain} 11.41 & \cellcolor{outdomain} 72.37 & \cellcolor{indomain} 12.21 \\
\hline
\end{tabular}
\caption{Results of jointly training networks with GT and PL.}
\label{table: dark knowledge as regularization} 
\end{table}

\textbf{Utlizing more accurate PL is crucial for satisfactory target-domain performance.} The diminishing difference between GT and PL correlates with an enhancement in target-domain performance. In particular, when employing GT in conjunction with PL($X_{c}(1)$), the network achieves a target-domain performance comparable to one trained solely with GT. It's worth noting that using the consistent region of PL($X_{c}(\tau)$) for regularization may not be optimal, as it can compromise domain generalization ability compared to using all regions of PL.

\begin{figure*}[t]
\centering
\includegraphics[width=1.0\linewidth]{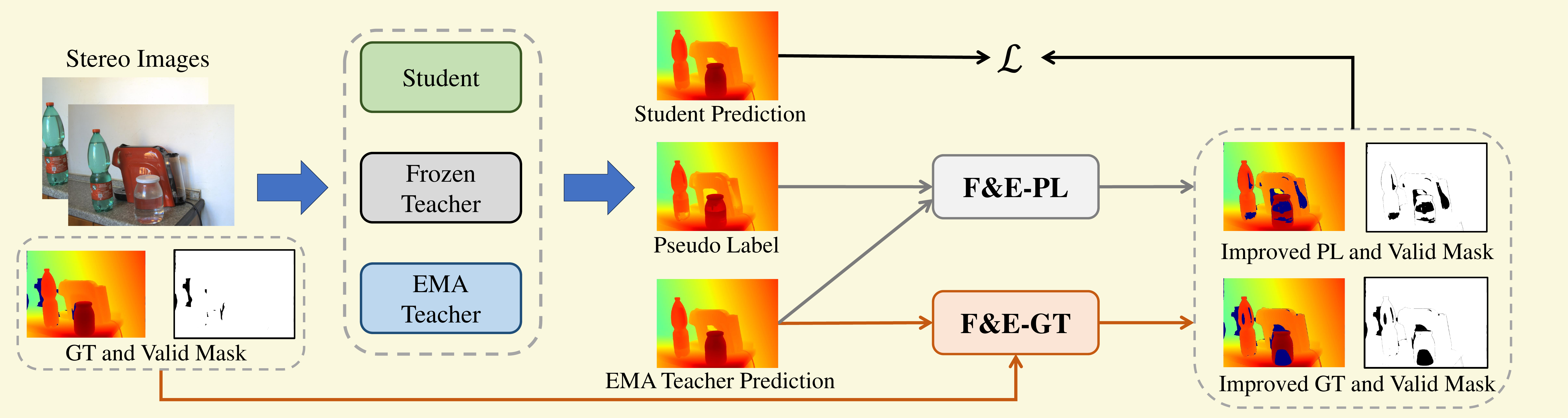}
\caption{Overview of the DKT framework. It uses the prediction from the EMA Teacher to improve GT and PL during fine-tuning.}
\label{fig: framework_overview}
\vspace{-2mm}
\end{figure*}

\textbf{More challenging scenarios degrade domain generalization ability more seriously.} The evaluation of the Booster dataset, characterized by numerous ToM surfaces, reveals a significant impact on degrading generalization ability. Through a comparison between fine-tuning networks on the Booster and KITTI datasets, we see the degradation on Bosster is more serious. Across various experimental settings, excluding the use of all PL regions, fine-tuning on Booster consistently results in an obvious decline in network robustness, emphasizing the importance of addressing the degradation posed by fine-tuning networks in such challenging scenarios for real-world applications.

\subsubsection{Summary}

Our investigation reveals that both GT and PL play dual roles during fine-tuning. GT proves crucial for improving target-domain performance, particularly in handling unseen challenges. However, insufficient regularization and overfitting GT details can degrade domain generalization ability. PL effectively preserves the generalization ability, however, learning incorrect predictions impact negatively. Moreover, a naive combination of GT and PL proves to be a conservative strategy that fails to leverage the full potential of both.

\section{DKT Framework}

\subsection{Architecture}

The core idea of DKT is to dynamically adjust the learning of different regions of GT and PL based on what networks have learned during fine-tuning. \Cref{fig: framework_overview} shows an overview of our framework.

DKT employs three same-initialized networks:
\begin{itemize}
    \item Student $\boldsymbol{\theta}_{S}$ is trained with GT and PL. After fine-tuning, we keep the Student for inference.
    \item Teacher $\boldsymbol{\theta}_{T}$ predicts PL and is frozen during fine-tuning. It contains the original knowledge from pre-training.
    \item EMA Teacher $\boldsymbol{\theta}_{T'}$ is an exponential moving average of Student. Here, it is used to measure what the Student has learned during fine-tuning.
\end{itemize}

The EMA Teacher is updated as a momentum-based moving average \cite{he2020momentum,tarvainen2017weight,athiwaratkun2018there} of the Student:
\begin{equation}
\setlength{\abovedisplayskip}{3pt}
\setlength{\belowdisplayskip}{3pt}
    \boldsymbol{\theta}_{T'} = m\boldsymbol{\theta}_{T'} + (1-m)\boldsymbol{\theta}_{S},
\end{equation}
where $m\in[0,1]$ is the momentum decay value for update speed control.

\begin{figure}[t]
\centering
\setlength{\abovecaptionskip}{3pt}
\includegraphics[width=1.0\linewidth]{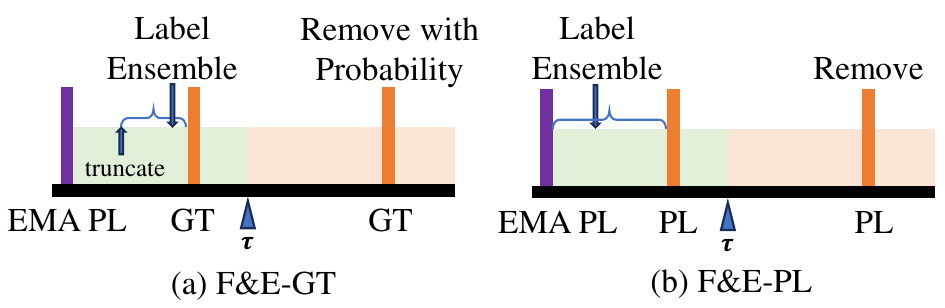}
\caption{The two variants of Filter and Ensemble (F\&E) applied to GT and PL. We use EMA PL to adjust different regions, with a threshold $\tau$ to control the operations on GT and PL.}
\label{fig: F&E}
\end{figure}

\begin{figure*}[t]
\centering
\setlength{\abovecaptionskip}{3pt}
\includegraphics[width=1.0\linewidth]{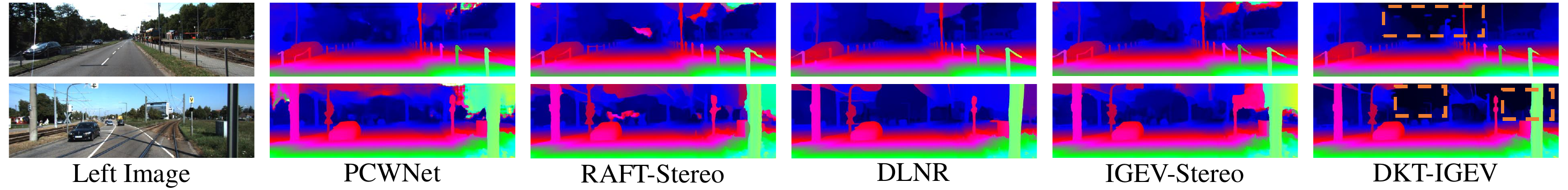}
\caption{Qualitative results on the KITTI benchmark.}
\label{fig: kitti benchmark}
\vspace{-2mm}
\end{figure*}

\begin{table*}[t]\footnotesize
\centering
\setlength{\tabcolsep}{5pt}
\begin{tabular}{c|ccccccccccccc}
\hline
\multirow{2}{*}{Method} & \multicolumn{2}{c}{KITTI 2012} & \multicolumn{3}{c}{KITTI 2015} & \multicolumn{5}{c}{DrivingStereo} & Middlebury & ETH3D & Booster \\
 & noc & all & bg & fg & all & sunny & cloudy & foggy & rainy & avg & \textgreater2px(\%)\  & \textgreater1px(\%)\ & \textgreater2px(\%)\ \\
\hline
\hline
PSMNet \cite{chang2018pyramid} & \cellcolor{indomain} 1.49 & \cellcolor{indomain} 1.89 & \cellcolor{indomain} 1.86 & \cellcolor{indomain} 4.62 & \cellcolor{indomain} 2.32 & \cellcolor{outdomain} 6.36 & \cellcolor{outdomain} 4.98 & \cellcolor{outdomain} 10.63 & \cellcolor{outdomain} 21.39 & \cellcolor{outdomain} 10.84 & \cellcolor{outdomain} 21.85 & \cellcolor{outdomain} 88.87 & \cellcolor{outdomain} 32.88 \\
GWCNet \cite{guo2019group} & \cellcolor{indomain} 1.32 & \cellcolor{indomain} 1.70 & \cellcolor{indomain} 1.74 & \cellcolor{indomain} 3.93 & \cellcolor{indomain} 2.11 & \cellcolor{outdomain} 3.52 & \cellcolor{outdomain} 2.84 & \cellcolor{outdomain} 4.27 & \cellcolor{outdomain} 9.04 & \cellcolor{outdomain} 4.92 & \cellcolor{outdomain} 17.83 & \cellcolor{outdomain} 31.87 & \cellcolor{outdomain} 29.08 \\
GANet \cite{zhang2019ga} & \cellcolor{indomain} 1.19 & \cellcolor{indomain} 1.60 & \cellcolor{indomain} 1.48 & \cellcolor{indomain} 3.46 & \cellcolor{indomain} 1.81 & \cellcolor{outdomain} 3.77 & \cellcolor{outdomain} 3.44 & \cellcolor{outdomain} 4.26 & \cellcolor{outdomain} 10.46 & \cellcolor{outdomain} 5.47 & \cellcolor{outdomain} 18.79 & \cellcolor{outdomain} 14.40 & \cellcolor{outdomain} 30.65 \\
PCWNet \cite{shen2022pcw} & \cellcolor{indomain} \underline{1.04} & \cellcolor{indomain} \underline{1.37} & \cellcolor{indomain} \underline{1.37} & \cellcolor{indomain} 3.16 & \cellcolor{indomain} 1.67 & \cellcolor{outdomain} 3.00 & \cellcolor{outdomain} 2.41 & \cellcolor{outdomain} \underline{1.72} & \cellcolor{outdomain} \underline{3.41} & \cellcolor{outdomain} 2.64 &  \cellcolor{outdomain} 15.55 & \cellcolor{outdomain} 18.27 & \cellcolor{outdomain} 21.34 \\
CGF-ACV \cite{xu2023cgi} & \cellcolor{indomain} \textbf{1.03} & \cellcolor{indomain} \textbf{1.34} & \cellcolor{indomain} \textbf{1.32} & \cellcolor{indomain} 3.08 & \cellcolor{indomain} \underline{1.61} & \cellcolor{outdomain} 3.22 & \cellcolor{outdomain} 3.19 & \cellcolor{outdomain} 6.69 & \cellcolor{outdomain} 17.50 & \cellcolor{outdomain} 7.65 &  \cellcolor{outdomain} 23.76  & \cellcolor{outdomain} 36.92 & \cellcolor{outdomain} 22.42 \\
RAFT-Stereo \cite{lipson2021raft} & \cellcolor{indomain} 1.30 & \cellcolor{indomain} 1.66 & \cellcolor{indomain} 1.58 & \cellcolor{indomain} 3.05 & \cellcolor{indomain} 1.82 & \cellcolor{outdomain} 2.18 & \cellcolor{outdomain} 1.91 & \cellcolor{outdomain} 2.74 & \cellcolor{outdomain} 8.35 & \cellcolor{outdomain} 3.80 & \cellcolor{outdomain} 11.78 & \cellcolor{outdomain} 40.43 & \cellcolor{outdomain} 23.87 \\
DLNR \cite{zhao2023high} & \cellcolor{indomain} - & \cellcolor{indomain} - & \cellcolor{indomain} 1.60 & \cellcolor{indomain} \underline{2.59} & \cellcolor{indomain} 1.76 & \cellcolor{outdomain} 1.94 & \cellcolor{outdomain} 1.87 & \cellcolor{outdomain} 2.25 & \cellcolor{outdomain} 5.31 & \cellcolor{outdomain} 2.84 & \cellcolor{outdomain} 8.21 & \cellcolor{outdomain} 26.18 & \cellcolor{outdomain} 19.58  \\
IGEV-Stereo \cite{xu2023iterative} & \cellcolor{indomain} 1.12 & \cellcolor{indomain} 1.44 & \cellcolor{indomain} 1.38 & \cellcolor{indomain} \textbf{2.67} & \cellcolor{indomain} \textbf{1.59} & \cellcolor{outdomain} 2.28 & \cellcolor{outdomain} 2.21 & \cellcolor{outdomain} 2.51 & \cellcolor{outdomain} 3.80 & \cellcolor{outdomain} 2.70 & \cellcolor{outdomain} 11.87 & \cellcolor{outdomain} 24.28 & \cellcolor{outdomain} 18.33 \\
DKT-RAFT(ours) & \cellcolor{indomain} 1.43 & \cellcolor{indomain} 1.85 & \cellcolor{indomain} 1.65 & \cellcolor{indomain} 2.98 & \cellcolor{indomain} 1.88 & \cellcolor{outdomain} \textbf{1.85} & \cellcolor{outdomain} \textbf{1.46} & \cellcolor{outdomain} \textbf{1.32} & \cellcolor{outdomain} 5.44 & \cellcolor{outdomain} \underline{2.52} &  \cellcolor{outdomain} \textbf{7.51} & \cellcolor{outdomain} \textbf{2.28} & \cellcolor{outdomain} \underline{15.35} \\
DKT-IGEV(ours) & \cellcolor{indomain} 1.22 & \cellcolor{indomain} 1.56 & \cellcolor{indomain} 1.46 & \cellcolor{indomain} 3.05 & \cellcolor{indomain} 1.72 & \cellcolor{outdomain} \underline{1.93} & \cellcolor{outdomain} \underline{1.71} & \cellcolor{outdomain} 1.96 & \cellcolor{outdomain} \textbf{3.26} & \cellcolor{outdomain} \textbf{2.22} & \cellcolor{outdomain} \underline{7.53} & \cellcolor{outdomain} \underline{4.23} & \cellcolor{outdomain} \textbf{15.30}  \\
\hline
\end{tabular}
\centering
\caption{Results on the KITTI benchmark. We evaluate the domain generalization ability with checkpoints provided by authors.}
\label{table: kitti benchmark}
\vspace{-2mm}
\end{table*}

We propose the \textbf{Filter and Ensemble (F\&E)} module, which utilizes the EMA Teacher to remove harmful regions and keep beneficial regions during fine-tuning. The division of different regions is conditioned on $\tau$, which is set to 3 in our framework. Based on the characteristics of GT and PL, it has two variants termed F\&E-GT and F\&E-PL. We present the two variants in \Cref{fig: F&E}.

\textbf{F\&E-GT.} For GT disparity map $D^{*}$, we remove the inconsistent region $X_{inc}(\tau)$ to avoid them dominating fine-tuning. However, since we hope networks learn to cope with new challenges, annotations of these challenging regions are especially valuable. Thus F\&E-GT removes the challenging inconsistent region with a probability based on the proportion in the whole valid region:
\begin{equation}
\begin{split}
\overline{D}^{*}_{inc} = \left \{
\begin{array}{ll}
D^{*}, & {\rm rand(0,1)} > \frac{|{X_{inc}(\tau)}|}{|X_{valid}|} \\
{\rm invalid}, & {\rm otherwise} \\
\end{array}
\right.
\end{split}
\end{equation}

It uses GT with a smaller probability for the more challenging region, thus avoiding new challenges disproportionately affecting learning. For the consistent region $X_{c}(\tau)$, we consider it an acceptable deviation and take continuous ensembling between GT $D^{*}$ and the EMA Teacher's prediction $\hat{D}^{T'}$. We sum them with uniform weights and truncate at 1 pixel apart from GT $D^{*}$:
\begin{equation}
    \alpha = {\rm rand(0,1)}
\end{equation}
\begin{equation}
    \overline{D}_{c}^{*} = \alpha * D^{*} + (1 - \alpha) * \hat{D}^{T'}.
\end{equation}
\begin{equation}
    \overline{D}^{*}_{c} = {\rm min}({\rm max}(\overline{D}_{c}^{*}, D^{*}-1), D^{*}+1).
\end{equation}

It adds fine-grained permutations to GT which prevents the Student from overfitting GT details. The final improved GT $\overline{D}^{*}$ is obtained by combining the improved inconsistent and consistent regions: 
\begin{equation}
\begin{split}
\overline{D}^{*}(x) = \left \{
\begin{array}{ll}
\overline{D}^{*}_{inc}(x), & x \in X_{inc}(\tau) \\
\overline{D}^{*}_{c}(x), & x \in X_{c}(\tau) \\
\end{array}
\right.
\end{split}
\end{equation}

\textbf{F\&E-PL.} The frozen Teacher predicts PL $\hat{D}^{T}$, serving as important regularization during fine-tuning. F\&E-PL aims to progressively enhance the accuracy of PL. We use the EMA Teacher's prediction $\hat{D}^{T'}$ to remove the inconsistent region of $\hat{D}^{T}$:
\begin{equation}
\setlength{\abovedisplayskip}{2pt}
    \hat{M} = |\hat{D}^{T} - \hat{D}^{T'}| < \tau.
\end{equation}
And we combine $\hat{D}^{T}$ and $\hat{D}^{T'}$ to get improved PL $\overline{D}^{T}$.
\begin{equation}
\setlength{\abovedisplayskip}{3pt}
\setlength{\belowdisplayskip}{2pt}
    \beta = {\rm rand(0,1)}
\end{equation}
\begin{equation}
    \overline{D}^{T} = \beta * \hat{D}^{T} + (1 - \beta) * \hat{D}^{T'}.
\end{equation}

\textbf{Training.} Our final training loss is a weighted sum of disparity loss with the improved GT $\overline{D}^{*}$ and PL $\overline{D}^{T}$ within valid masks $M^{*}$ for GT and $\hat{M}$ for PL:  
\begin{equation}
    \mathcal{L} = \mathcal{L}^{disp}(\hat{D}, \overline{D}^{*}, M^{*}) + \lambda \mathcal{L}^{disp}(\hat{D}, \overline{D}^{T}, \hat{M}),
\end{equation}
where $\hat{D}$ is the Student prediction and $\lambda$ is the balancing weight. After 5k steps of fine-tuning, we re-initialize the EMA Teacher with the Student weights. The re-initialized EMA Teacher predicts more accurate disparities, resulting in more accurate augmented GT and invalid-region PL.

\begin{figure}[t]
\centering
\setlength{\abovecaptionskip}{5pt}
\includegraphics[width=1.0\linewidth]{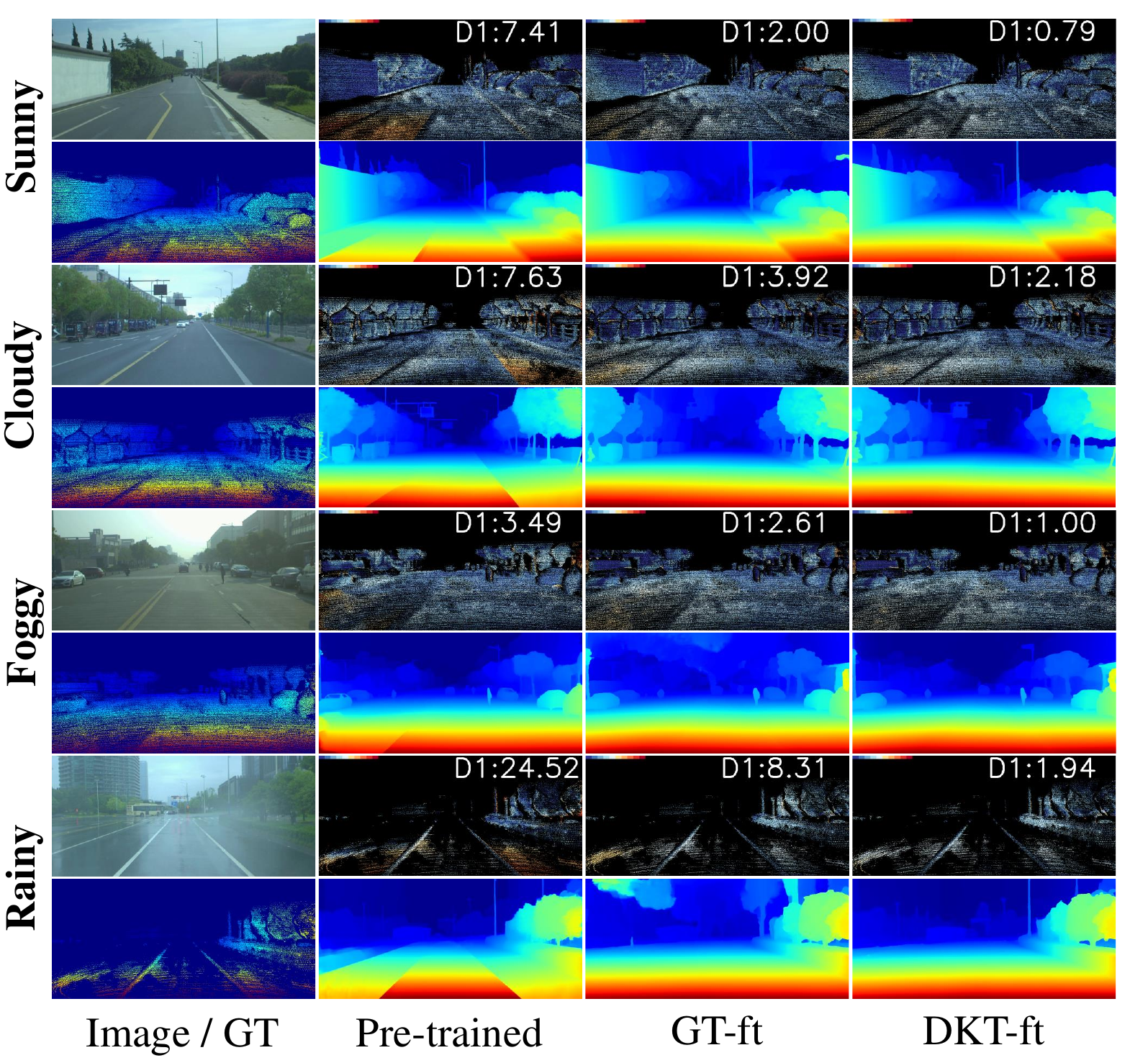}
\caption{Qualitative results on the DrivingStereo dataset. We fine-tune the networks with KITTI datasets.}
\label{fig: driving stereo}
\end{figure}

\begin{figure*}
  \begin{minipage}{0.7\textwidth}
    \captionsetup{type=table}
    \footnotesize
    \centering
    \begin{tabular}{cccc|ccccc}
    \hline
    $\mathcal{L}_{GT}$ & $\mathcal{L}_{PL}$ & F\&E-GT & F\&E-PL  & 2012 & 2015 & Midd & ETH3D & Booster \\
    \hline
    \hline
    Training set & \multicolumn{8}{c}{\textbf{KITTI 2012 \& 2015}} \\
    \hline
    \textcolor{Dgreen}{\cmark} & &  & & \cellcolor{indomain} 1.94 & \cellcolor{indomain} 1.36 & \cellcolor{outdomain} 12.23 & \cellcolor{outdomain} 23.88 & \cellcolor{outdomain} 18.43 \\
    \textcolor{Dgreen}{\cmark} & &  \textcolor{Dgreen}{\cmark} & & \cellcolor{indomain} 1.93 & \cellcolor{indomain} 1.38 & \cellcolor{outdomain} 9.62 & \cellcolor{outdomain} 12.31 & \cellcolor{outdomain} 17.46 \\
    \textcolor{Dgreen}{\cmark} & \textcolor{Dgreen}{\cmark} & & & \cellcolor{indomain} 3.24 & \cellcolor{indomain} 2.75 & \cellcolor{outdomain} 7.40 & \cellcolor{outdomain} 3.64 & \cellcolor{outdomain} 16.54 \\
    \textcolor{Dgreen}{\cmark} & \textcolor{Dgreen}{\cmark} & \textcolor{Dgreen}{\cmark} & & \cellcolor{indomain} 3.26 & \cellcolor{indomain} 2.79 & \cellcolor{outdomain} 7.03 & \cellcolor{outdomain} 3.24 & \cellcolor{outdomain} 15.19 \\
    \textcolor{Dgreen}{\cmark} & \textcolor{Dgreen}{\cmark} &  & \textcolor{Dgreen}{\cmark} & \cellcolor{indomain} 1.97 & \cellcolor{indomain} 1.43 & \cellcolor{outdomain} 8.08 & \cellcolor{outdomain} 4.23 & \cellcolor{outdomain} 15.11 \\
    \textcolor{Dgreen}{\cmark} & \textcolor{Dgreen}{\cmark} & \textcolor{Dgreen}{\cmark} & \textcolor{Dgreen}{\cmark} & \cellcolor{indomain} 1.98 & \cellcolor{indomain} 1.39 & \cellcolor{outdomain} 7.11 & \cellcolor{outdomain} 3.64 & \cellcolor{outdomain} 15.51 \\
    \hline
    \hline
    Training set & \multicolumn{8}{c}{\textbf{Booster}} \\
    \hline
    \textcolor{Dgreen}{\cmark} & &  & & \cellcolor{outdomain} 52.30 & \cellcolor{outdomain} 55.44 & \cellcolor{outdomain} 19.78 & \cellcolor{outdomain} 93.31 & \cellcolor{indomain} 12.88 \\
    \textcolor{Dgreen}{\cmark} & &  \textcolor{Dgreen}{\cmark} & &  \cellcolor{outdomain} 9.42 & \cellcolor{outdomain} 11.04 & \cellcolor{outdomain} 7.56 & \cellcolor{outdomain} 6.11 & \cellcolor{indomain} 12.76 \\
    \textcolor{Dgreen}{\cmark} & \textcolor{Dgreen}{\cmark} &  & &  \cellcolor{outdomain} 7.14 & \cellcolor{outdomain} 10.69 & \cellcolor{outdomain} 8.02 & \cellcolor{outdomain} 10.30 & \cellcolor{indomain} 14.29 \\
    \textcolor{Dgreen}{\cmark} & \textcolor{Dgreen}{\cmark} &  \textcolor{Dgreen}{\cmark} & & \cellcolor{outdomain} 3.69 & \cellcolor{outdomain} 4.88 & \cellcolor{outdomain} 7.01 & \cellcolor{outdomain} 4.32 & \cellcolor{indomain} 14.21 \\
    \textcolor{Dgreen}{\cmark} & \textcolor{Dgreen}{\cmark} &  & \textcolor{Dgreen}{\cmark} & \cellcolor{outdomain} 23.17 & \cellcolor{outdomain} 24.58 & \cellcolor{outdomain} 10.56 & \cellcolor{outdomain} 47.91 & \cellcolor{indomain} 12.55 \\
    \textcolor{Dgreen}{\cmark} & \textcolor{Dgreen}{\cmark} &  \textcolor{Dgreen}{\cmark} & \textcolor{Dgreen}{\cmark} &  \cellcolor{outdomain} 3.49 & \cellcolor{outdomain} 4.71 & \cellcolor{outdomain} 6.61 & \cellcolor{outdomain} 2.60 & \cellcolor{indomain} 12.63 \\
    \hline
    \end{tabular}
    \caption{Ablation study of each component of DKT framework on the KITTI 2012 and 2015, Middlebury, ETH3D, and Booster training sets.}
    \label{table: ablation of key components}
  \end{minipage}
  \hspace{0.02\textwidth}
  \begin{minipage}{0.25\textwidth}
    \captionsetup{type=figure}
    \centering
    \includegraphics[width=0.905\linewidth]{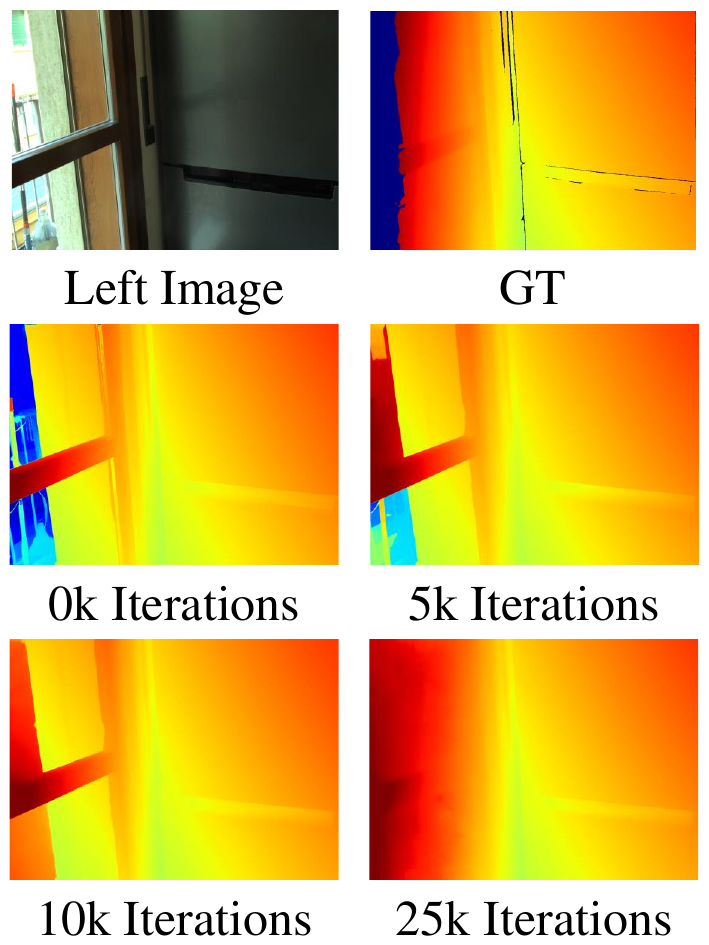}
    \caption{Predictions of the EMA Teacher during fine-tuning.}
    \label{fig: EMA Teacher changes}
  \end{minipage}
\end{figure*}

\subsection{Overall Performance}

\subsubsection{KITTI Benchmark}

We compare the performance of DKT and published SOTA methods \cite{shen2022pcw,xu2023cgi,zhao2023high,xu2023iterative} on the KITTI benchmark by considering target-domain  and cross-domain performance. We utilize checkpoints provided by the authors to assess the performance of domain generalization. For methods to submit separate checkpoints on 2012 and 2015 benchmarks, we calculate the average of the two checkpoints. \Cref{table: kitti benchmark} shows the results. For target KITTI scenarios, we find our methods can achieve good but slightly worse performance than the base networks we use. We show the qualitative results in \Cref{fig: kitti benchmark}, compared to other methods, DKT demonstrates reasonable predictions for weak texture regions. We evaluate how these models generalize to DrivingStereo, which contains unseen driving scenarios with challenging weathers. Among these published models, DKT achieves the strongest robustness by considering four challenging weathers. We visualize results in \Cref{fig: driving stereo}. We evaluate if these models can generalize well to Middlebury, ETH3D, and Booster datasets. We see that DKT is the only method that can generalize well to these scenarios after fine-tuning.

\begin{table}[t]\footnotesize
\centering
\setlength{\tabcolsep}{4.5pt}
\begin{tabular}{c|ccccc}
\hline
Method & Booster & 2012 & 2015 & Midd & ETH3D \\
\hline
\hline
CFNet \cite{shen2021cfnet} * & 38.32 & 4.97 &  6.31 &  15.77 &  5.47 \\
RAFT-Stereo \cite{lipson2021raft} * & 17.44 & 4.34 &  5.68 &  8.41 &  \textbf{2.29} \\
\hdashline
CFNet(ft) \cite{shen2021cfnet} \dag & \cellcolor{indomain} 29.65 & \cellcolor{outdomain} 51.52 & \cellcolor{outdomain} 68.24 & \cellcolor{outdomain} 18.42 & \cellcolor{outdomain} 79.98 \\
RAFT-Stereo(ft) \cite{lipson2021raft} \dag & \cellcolor{indomain} 10.73 & \cellcolor{outdomain} 17.02 & \cellcolor{outdomain} 23.33 & \cellcolor{outdomain} 15.98 & \cellcolor{outdomain} 89.05 \\
PCVNet(ft) \cite{zeng2023parameterized}    \ddag  & \cellcolor{indomain} \textbf{9.03} & \cellcolor{outdomain} 35.54 & \cellcolor{outdomain} 38.60 & \cellcolor{outdomain} 22.88 & \cellcolor{outdomain} 75.21 \\
DKT-RAFT(ours) & \cellcolor{indomain} 10.32  & \cellcolor{outdomain} \textbf{3.69} &  \cellcolor{outdomain} 4.95 & \cellcolor{outdomain} \textbf{5.94}  & \cellcolor{outdomain} 2.57 \\
DKT-IGEV(ours) & \cellcolor{indomain} 14.11  & \cellcolor{outdomain} 3.76 &  \cellcolor{outdomain} \textbf{4.78} & \cellcolor{outdomain} 6.94  & \cellcolor{outdomain} 3.13 \\
\hline
\end{tabular}
\centering
\caption{Results on Booster benchmark. For target-domain performance, we report their online results. For generalization evaluation: * uses authors' weights, \dag reproduces models in the same setting as online submissions, \ddag the original implementation for submission uses CREStereo dataset \cite{li2022practical} to augment Booster and we retrain it with only Booster training set to avoid unfair comparison.}
\label{table: booster benchmark}
\vspace{-2mm}
\end{table}

\subsubsection{Booster Benchmark}

We follow \cite{ramirez2022open,costanzino2023learning,zeng2023parameterized} to fine-tune networks on the Booster dataset, which contains very challenging real-world scenarios. The results are presented in \Cref{table: booster benchmark}. We see fine-tuning networks with GT improves the predictions in the target domain compared to the synthetic data pre-trained networks. However, it also introduces a notable degradation in generalization ability when fine-tuning networks in such challenging scenarios. Using our DKT framework to fine-tune networks achieves competitive target-domain performances compared to the baseline fine-tuning. Moreover, DKT can help networks generalize knowledge learned from the challenging Booster data to unseen real-world scenes, improving the performance on KITTI and Middlebury than the synthetic data pre-trained models.

\subsection{Ablation Study}
We evaluate the effectiveness of DKT in utilizing the EMA Teacher for improving PL and GT in \Cref{table: ablation of key components}. Additionally, we provide visual insights into the EMA Teacher progressively predicts more accurate disparities in \Cref{fig: EMA Teacher changes}.

\textbf{F\&E-GT.} It improves GT by combining GT and the prediction of the EMA Teacher. It removes the inconsistent region of GT with certain probabilities and adds fine-grained permutations to GT. It also works when we only use GT supervision $\mathcal{L}_{GT}$ and do not introduce PL supervision $\mathcal{L}_{PL}$: only using F\&E-GT to improve GT alleviates the domain generalization ability degradation and little impacts the target-domain performances. For using both PL supervision $\mathcal{L}_{PL}$ and F\&E-GT, we show this combination can effectively preserve the domain generalization ability. However, using $\mathcal{L}_{PL}$ negatively impacts the target-domain performances because $\mathcal{L}_{PL}$ can introduce incorrect supervision, which we solve with F\&E-PL.

\textbf{F\&E-PL.} It plays an important role in improving target-domain performances by removing the inconsistent region of PL. It progressively identifies incorrect disparities of PL according to EMA Teacher's prediction, instead of directly comparing PL with GT which can degrade the domain generalization ability in \Cref{sec: dark knowledge as regularization with GT}. We show that only F\&E-PL can alleviate the domain generalization degradation in KITTI, however, it cannot work well alone in the challenging Booster scenarios where using F\&E-GT is necessary.

\section{Conclusion}
We aim to fine-tune stereo networks without compromising robustness to unseen domains. We identify that learning new knowledge without sufficient regularization and overfitting GT details can degrade the robustness. We propose the DKT framework, which improves fine-tuning by dynamically measuring what has been learned. Experiments show that DKT can preserve robustness during fine-tuning, improve robustness to challenging weather, and generalize knowledge learned from target domains to unseen domains.

\vspace{2pt}\noindent\textbf{Acknowledgements.}
This work was supported by the National Science and Technology Major Project under Grant 2022ZD0116310, the National Natural Science Foundation of China 62276016, 62372029.

\clearpage
\maketitlesupplementary

\setcounter{section}{0}
\setcounter{figure}{0}
\setcounter{table}{0}

\renewcommand\thesection{\Alph{section}}
\renewcommand\thefigure{\Roman{figure}}
\renewcommand\thetable{\Roman{table}}

\section*{Overview}

We organize the material as follows. \Cref{supp_sec: experimental Details} shows more details of the experiment settings. In \Cref{supp_sec: network architectures for GT vs. PL}, we provide comparison results between using GT and PL for fine-tuning with the additional stereo matching network architecture. \Cref{supp_sec: more ablations about DKT} and \Cref{supp_sec: results of more experimental settgins} conducts additional experiments about DKT. \Cref{supp_sec: more visualization results} presents more qualitative results of the domain generalization performance.

\section{Details of Experimental Setting} \label{supp_sec: experimental Details}

\noindent \textbf{Dataset.} We conduct our experiments by initializing stereo matching networks with synthetic dataset pre-trained weights and fine-tuning them in real-world scenarios. We mainly focus on the robustness and domain generalization ability after fine-tuning networks, and we also show their target-domain performances to ensure networks actually learn from target domains. When not specifically mentioned, all networks in our experiments are pre-trained on the SceneFlow \cite{mayer2016large} dataset. The introduction of the synthetic and real-world datasets are as follows:
\begin{itemize}
\item \textit{SceneFlow} \cite{mayer2016large} is a large synthetic dataset that consists of 35,454 pairs of stereo images for training and 4,370 pairs for evaluation. Both sets have dense ground-truth disparities. The resolution of the images is $960\times540$. Besides the original clean pass, the dataset also contains a final pass. The final pass has motion blur and defocus blur, making it more similar to real-world images. SceneFlow is currently the most commonly used dataset for pre-training stereo matching networks.
\item \textit{KITTI 2012} \cite{geiger2012we} collects outdoor driving scenes with sparse ground-truth disparities. It contains 194 training samples and 195 testing samples with a resolution of $1226 \times 370$.
\item \textit{KITTI 2015} \cite{menze2015object} collects driving scenes with sparse disparity maps. It contains 200 training samples and 200 testing samples with a resolution of $1242 \times 375$.
\item \textit{Booster} \cite{ramirez2022open} contains 228 samples for training and 191 samples for online testing in 64 different scenes with dense ground-truth disparities. Most of the collected scenes have challenging non-Lambertian surfaces. We use the quarter resolution in our experiments.
\item \textit{Middlebury} \cite{scharstein2014high} consists of 15 training and 15 testing stereo pairs captured indoors. The dataset offers images at full, half, and quarter resolutions. We use the half-resolution training set for domain generalization evaluation.
\item \textit{ETH3D} \cite{schops2017multi} consists of 27 grayscale image pairs for training and 20 for testing. It includes both indoor and outdoor scenes. We use the training set for domain generalization evaluation.
\item \textit{DrivingStereo} \cite{yang2019drivingstereo} is a large-scale real-world driving dataset. A subset of it contains 2,000 stereo pairs collected under different weather (sunny, cloudy, foggy, and rainy). We use the half resolution of these challenging scenes to evaluate the robustness after fine-tuning on the KITTI datasets.
\end{itemize}

\noindent \textbf{Local Dataset Split.} Except for online submissions, we conduct the experiments based on local train and validation splits. For the KITTI 2012 and 2015 datasets, we follow GWCNet \cite{guo2019group} to split 14 stereo pairs of 2012 and 20 pairs of 2015 for validation, the remaining 360 pairs are used for training. For the booster dataset, we use the `Washer' and `OilCan' scenes (15 stereo pairs) for validation, and the remaining 213 pairs for training. In this material, we also conduct fine-tuning experiments on Middlebury and ETH3D datasets, following the data split in \cite{liang2019stereo}. We use the `ArtL' and `Playroom' scenes (2 stereo pairs) for Middlebury validation and the `facade' and `forest' scenes (3 stereo pairs) for ETH3D validation.

\section{Network Architecture for GT vs. PL} \label{supp_sec: network architectures for GT vs. PL}

In the main paper, we investigate the distinct behaviors of Ground Truth (GT) and Pseudo Label (PL) during fine-tuning. We achieve this by dividing pixels into different regions ($X_{c}(\tau)$, $X_{inc}(\tau)$, $X_{invalid}$) and conducting comprehensive comparisons between them. In addition to the iterative optimization based IGEV-Stereo \cite{xu2023iterative}, we employ the 3D convolution-based CFNet \cite{shen2021cfnet} to affirm that our findings are applicable across diverse stereo matching network architectures. As presented in \Cref{supp_table: explore dark knowledge with CFNet}, learning new knowledge without sufficient regularization and overfitting GT details are two primary contributors to the degradation of domain generalization ability during the fine-tuning.

\begin{table}[h]\footnotesize
\setlength{\abovecaptionskip}{6pt}
\centering
\begin{tabular}{c|ccccc}
\hline
Supervision & 2012 & 2015 & Midd & ETH3D & Booster \\
\hline
\hline
Training set & \multicolumn{5}{c}{\textbf{KITTI 2012 \& 2015}} \\
\hline
zero-shot & 5.71 & 4.84 & 15.77 & 5.48 & 38.84 \\
\hdashline 
GT(valid) & \cellcolor{indomain} \textbf{2.15} & \cellcolor{indomain} \textbf{1.39} & \cellcolor{outdomain} 19.83 & \cellcolor{outdomain} 29.94 & \cellcolor{outdomain} 30.95 \\
GT($X_{c}(3)$) & \cellcolor{indomain} 2.26 & \cellcolor{indomain} 1.67 & \cellcolor{outdomain} 17.92 & \cellcolor{outdomain} 24.52 & \cellcolor{outdomain} 30.88 \\
GT($X_{inc}(3)$) & \cellcolor{indomain} 21.33 & \cellcolor{indomain} 18.49 & \cellcolor{outdomain} 31.78 & \cellcolor{outdomain} 58.35 & \cellcolor{outdomain} 43.06 \\
GT($X_{c}(1)$) & \cellcolor{indomain} 2.67 & \cellcolor{indomain} 1.95 & \cellcolor{outdomain} 16.27 & \cellcolor{outdomain} 14.67 & \cellcolor{outdomain} 31.16 \\
\hdashline
PL(all) & \cellcolor{indomain} 5.05 & \cellcolor{indomain} 4.26 & \cellcolor{outdomain} \textbf{13.71} & \cellcolor{outdomain} \textbf{4.86} & \cellcolor{outdomain} \textbf{29.92} \\
PL(valid) & \cellcolor{indomain} 5.58 & \cellcolor{indomain} 4.64 & \cellcolor{outdomain} 14.78 & \cellcolor{outdomain} 6.05 & \cellcolor{outdomain} 30.79 \\
PL($X_{c}(3)$) & \cellcolor{indomain} 3.32 & \cellcolor{indomain} 3.01 & \cellcolor{outdomain} 14.09 & \cellcolor{outdomain} 5.50 & \cellcolor{outdomain} 30.97 \\
PL($X_{inc}(3)$) & \cellcolor{indomain} 8.91 & \cellcolor{indomain} 8.08 & \cellcolor{outdomain} 18.30 & \cellcolor{outdomain} 12.45 & \cellcolor{outdomain} 40.17 \\
PL($X_{c}(1)$) & \cellcolor{indomain} 2.94 & \cellcolor{indomain} 2.57 & \cellcolor{outdomain} 15.38 & \cellcolor{outdomain} 5.80 & \cellcolor{outdomain} 31.34 \\
\hline
Training set & \multicolumn{5}{c}{\textbf{Booster}} \\
\hline
zero-shot & 4.97 & 6.31 & 15.77 & 5.48 & 35.03 \\
\hdashline 
GT(valid) & \cellcolor{outdomain} 56.20 & \cellcolor{outdomain} 71.41 & \cellcolor{outdomain} 18.45 & \cellcolor{outdomain} 80.53 & \cellcolor{indomain} \textbf{25.86} \\
GT($X_{c}(3)$) & \cellcolor{outdomain} 4.39 & \cellcolor{outdomain} 6.04 & \cellcolor{outdomain} 13.53 & \cellcolor{outdomain} 20.90 & \cellcolor{indomain} 26.85 \\
GT($X_{inc}(3)$) & \cellcolor{outdomain} 97.27 & \cellcolor{outdomain} 97.86 & \cellcolor{outdomain} 77.44 & \cellcolor{outdomain} 99.89 & \cellcolor{indomain} 45.69 \\
GT($X_{c}(1)$) & \cellcolor{outdomain} 4.31 & \cellcolor{outdomain} 6.19 & \cellcolor{outdomain} 13.77 & \cellcolor{outdomain} 21.80 & \cellcolor{indomain} 26.13 \\
\hdashline
PL(all) & \cellcolor{outdomain} 4.24 & \cellcolor{outdomain} 5.19 & \cellcolor{outdomain} 11.38 & \cellcolor{outdomain} 5.42 & \cellcolor{indomain} 28.34 \\
PL(valid) & \cellcolor{outdomain} 4.33 & \cellcolor{outdomain} 5.11 & \cellcolor{outdomain} 11.48 & \cellcolor{outdomain} 5.51 & \cellcolor{indomain} 28.05 \\
PL($X_{c}(3)$) & \cellcolor{outdomain} 3.98 & \cellcolor{outdomain} \textbf{4.65} & \cellcolor{outdomain} 11.25 & \cellcolor{outdomain} \textbf{5.39} & \cellcolor{indomain} 27.40 \\
PL($X_{inc}(3)$) & \cellcolor{outdomain} 5.56 & \cellcolor{outdomain} 7.68 & \cellcolor{outdomain} 16.81 & \cellcolor{outdomain} 6.08 & \cellcolor{indomain} 36.44 \\
PL($X_{c}(1)$) & \cellcolor{outdomain} \textbf{3.79} & \cellcolor{outdomain} 4.98 & \cellcolor{outdomain} \textbf{11.03} & \cellcolor{outdomain} 5.59 & \cellcolor{indomain} 27.54 \\
\hline
\end{tabular}
\caption{Results of using different regions of GT or PL to fine-tune CFNet \cite{shen2021cfnet}. Different regions play varied roles during fine-tuning.}
\label{supp_table: explore dark knowledge with CFNet} 
\end{table}

\section{Additional Ablations about DKT} \label{supp_sec: more ablations about DKT}

\subsection{Fine-grained Permutations}

\begin{figure}[t]
    \setlength{\abovecaptionskip}{3pt} 
    \centering
    \includegraphics[width=1\linewidth]{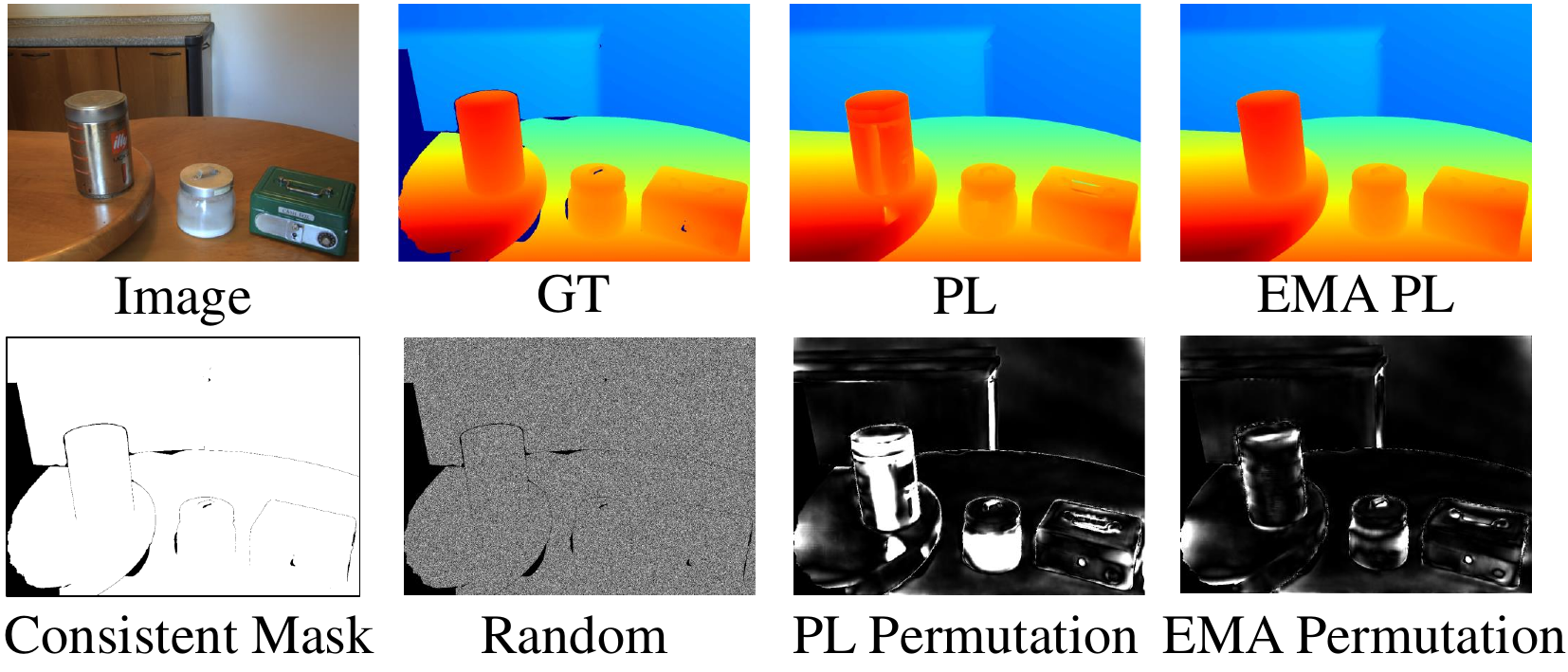}
    \caption{Visualization with the absolute value of three kinds of fine-grained permutations.}
    \label{suppl_fig: permutations}
    \vspace{-2mm}
\end{figure}

In F\&E-GT, we leverage the exponential moving average (EMA) Teacher's prediction to serve as fine-grained permutations for GT. In this section, we present ablations with alternative permutations. Specifically, we apply F\&E-GT using the EMA Teacher for filtering out inconsistent regions but with variations in permutations. We use random noise from (-1, 1), PL from the frozen Teacehr, and the EMA Teacher's prediction for ablation and visualize the three kinds of fine-grained permutations in \Cref{suppl_fig: permutations}. The results are presented in \Cref{supp_table: fine-grained permutations}. Our findings demonstrate that employing the frozen Teacher or EMA Teacher to add permutations better preserves domain generalization ability than random noise. Moreover, utilizing the EMA Teacher yields better target-domain performance compared to the frozen Teacher. We attribute this improvement to the EMA Teacher progressively predicting more accurate disparities.

\begin{table}[h]\footnotesize
\setlength{\abovecaptionskip}{6pt}
\centering
\begin{tabular}{c|ccccc}
\hline
Method & 2012 & 2015 & Midd & ETH3D & Booster \\
\hline
\hline
Training set & \multicolumn{5}{c}{\textbf{KITTI 2012 \& 2015}} \\
\hline
random noise & \cellcolor{indomain} 1.94 & \cellcolor{indomain} 1.39 &  \cellcolor{outdomain} 11.08 & \cellcolor{outdomain} 19.79  & \cellcolor{outdomain} 17.93 \\
F.T. & \cellcolor{indomain} 1.94 & \cellcolor{indomain} 1.40 &  \cellcolor{outdomain} \textbf{9.60} & \cellcolor{outdomain} 12.34  & \cellcolor{outdomain} 17.57 \\
EMA.T. & \cellcolor{indomain} \textbf{1.93} & \cellcolor{indomain} \textbf{1.38} & \cellcolor{outdomain} 9.62 & \cellcolor{outdomain} \textbf{12.31} & \cellcolor{outdomain} \textbf{17.46} \\
\hline
\hline
Training set  & \multicolumn{5}{c}{\textbf{Booster}} \\
\hline
random noise & \cellcolor{outdomain} 11.55 & \cellcolor{outdomain} 12.28 &  \cellcolor{outdomain} 9.54 & \cellcolor{outdomain} 7.73  & \cellcolor{indomain} 12.85 \\
F.T. & \cellcolor{outdomain} 9.48 & \cellcolor{outdomain} \textbf{10.96} &  \cellcolor{outdomain} 7.81 & \cellcolor{outdomain} \textbf{5.93 } & \cellcolor{indomain} 12.89 \\
EMA.T. &  \cellcolor{outdomain} \textbf{9.4}2 & \cellcolor{outdomain} 11.04 & \cellcolor{outdomain}\textbf{ 7.56} & \cellcolor{outdomain} 6.11 & \cellcolor{indomain} \textbf{12.76} \\
\hline
\end{tabular}
\centering
\caption{Ablation of fine-grained permutations. F.T.: the frozen Teacher. PL serves as better permutations than random noise.}
\label{supp_table: fine-grained permutations} 
\vspace{-2mm}
\end{table}

\subsection{Effects of the Frozen Teacher}

\begin{figure}[h]
    \setlength{\abovecaptionskip}{3pt} 
    \centering
    \includegraphics[width=1\linewidth]{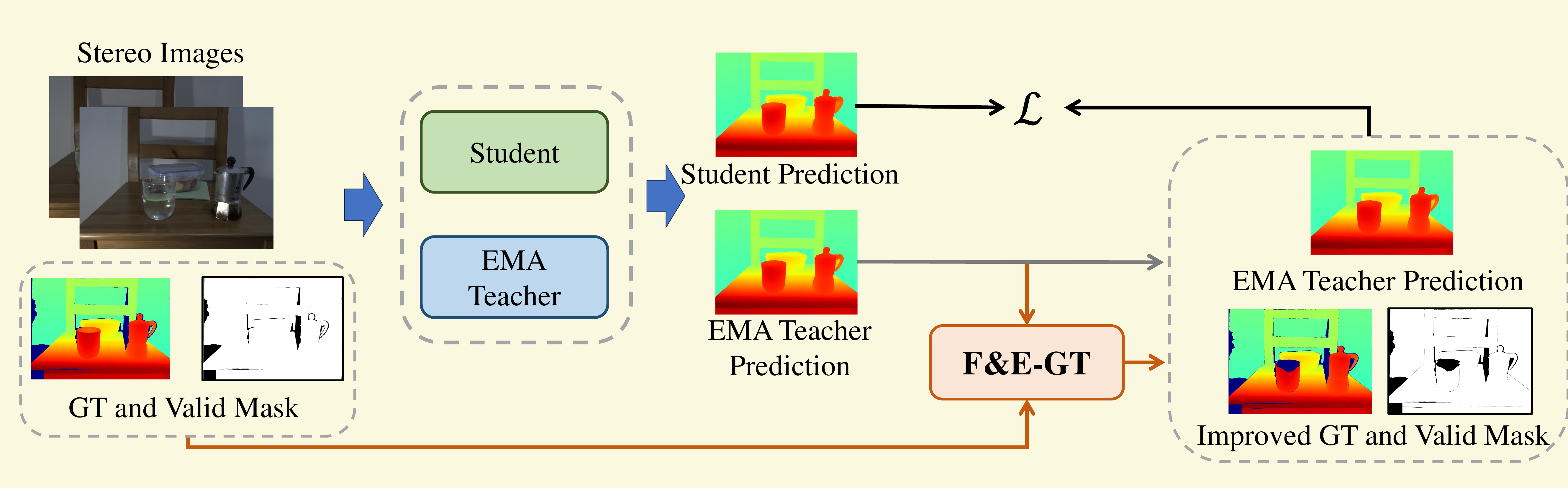}
    \caption{DKT framework without the frozen Teacher.}
    \label{suppl_fig: dkt wo frozen Teacher}
    \vspace{-2mm}
\end{figure}

An alternative way to using the frozen Teacher's prediction with F\&E-PL is directly using the EMA Teacher's prediction, which progressively predicts more accurate disparities. An overview of DKT without the frozen Teacher is shown in \Cref{suppl_fig: dkt wo frozen Teacher}. We show the comparison in \Cref{supp_table: effects of frozen and ema}. Using the frozen Teacher's prediction gets improvements in target domains than using the frozen Teacher's prediction with F\&E-PL, however, it leads to a slight drop in domain generalization ability than using the Frozen Teacher's prediction.

\begin{table}[h]\footnotesize
\setlength{\abovecaptionskip}{6pt}
\centering
\begin{tabular}{c|ccccc}
\hline
Method & 2012 & 2015 & Midd & ETH3D & Booster \\
\hline
\hline
Training set & \multicolumn{5}{c}{\textbf{KITTI 2012 \& 2015}} \\
\hline
DKT(w/o F.T.) & \cellcolor{indomain} \textbf{1.97}  & \cellcolor{indomain} \textbf{1.34} &  \cellcolor{outdomain} 8.02 & \cellcolor{outdomain} 4.43  & \cellcolor{outdomain} 15.65 \\
DKT(full) & \cellcolor{indomain} 1.98  & \cellcolor{indomain} 1.39 &  \cellcolor{outdomain} \textbf{7.11} & \cellcolor{outdomain} \textbf{3.64}  & \cellcolor{outdomain} \textbf{15.51} \\
\hline
\hline
Training set  & \multicolumn{5}{c}{\textbf{Booster}} \\
\hline
DKT(w/o F.T.) & \cellcolor{outdomain} 3.72  & \cellcolor{outdomain} 4.86 & \cellcolor{outdomain} 7.00 & \cellcolor{outdomain} 3.89 & \cellcolor{indomain} \textbf{12.23} \\
DKT(full) & \cellcolor{outdomain} \textbf{3.49}  & \cellcolor{outdomain} \textbf{4.71} & \cellcolor{outdomain} \textbf{6.61}& \cellcolor{outdomain} \textbf{2.60} & \cellcolor{indomain} 12.63 \\
\hline
\end{tabular}
\centering
\caption{Effects of the using frozen Teacher to produce PL. F.T.: the frozen Teacher. Using the frozen Teacher to produce PL preserves the domain generalization ability better.}
\label{supp_table: effects of frozen and ema}
\vspace{-2mm}
\end{table}

\section{Additional Experiments about DKT} \label{supp_sec: results of more experimental settgins}

\subsection{DKT vs. DG Methods} \label{supp_sec: comparison with dg methods}

Training robust stereo matching networks on synthetic datasets has been well-researched recently \cite{zhang2020domain,zhang2022revisiting,chuah2022itsa}. We compare our methods with domain generalization methods and verify if these methods work well for real-world fine-tuning. We use ITSA \cite{chuah2022itsa} and Asymmetric Augmentation \cite{yang2019hierarchical} for comparison. As shown in \Cref{table: comparison with generalization methods},  domain generalization methods designed for synthetic data pre-training fail in this case. We think differences between synthetic and real data may render previous methods unsuitable: existing methods reduce shortcuts learning caused by synthetic artifacts \cite{chuah2022itsa}, while real-world data is actually real and the factors degrade generalization ability can be different.

\begin{table}[h]\footnotesize
\setlength{\abovecaptionskip}{6pt}
\centering
\begin{tabular}{c|ccccc}
\hline
Method & 2012 & 2015 & Midd & ETH3D & Booster \\
\hline
\hline
Training set & \multicolumn{5}{c}{\textbf{KITTI 2012 \& 2015}} \\
\hline
baseline & \cellcolor{indomain} \textbf{1.94} & \cellcolor{indomain} 1.36 & \cellcolor{outdomain} 12.23 & \cellcolor{outdomain} 23.88 & \cellcolor{outdomain} 18.43 \\
ITSA \cite{chuah2022itsa} & \cellcolor{indomain} 2.01 & \cellcolor{indomain} 1.43 & \cellcolor{outdomain} 12.59 & \cellcolor{outdomain} 25.72 & \cellcolor{outdomain} 17.98 \\
Asy.Aug  & \cellcolor{indomain} 1.98 & \cellcolor{indomain} \textbf{1.34} & \cellcolor{outdomain} 12.67  & \cellcolor{outdomain} 23.37 & \cellcolor{outdomain} 18.02 \\
DKT & \cellcolor{indomain} 1.98  & \cellcolor{indomain} 1.39 &  \cellcolor{outdomain} \textbf{7.11} & \cellcolor{outdomain} \textbf{3.64}  & \cellcolor{outdomain} \textbf{15.51} \\
\hline
\hline
Training set  & \multicolumn{5}{c}{\textbf{Booster}} \\
\hline
baseline   & \cellcolor{outdomain} 52.30  & \cellcolor{outdomain} 55.44  & \cellcolor{outdomain} 19.78  & \cellcolor{outdomain} 93.31  & \cellcolor{indomain} 12.88 \\
ITSA \cite{chuah2022itsa} & \cellcolor{outdomain} 51.95  & \cellcolor{outdomain} 56.97 & \cellcolor{outdomain} 18.77 & \cellcolor{outdomain} 98.78 & \cellcolor{indomain} 13.01 \\
Asy.Aug  & \cellcolor{outdomain} 55.79  & \cellcolor{outdomain} 58.36 & \cellcolor{outdomain} 18.51 & \cellcolor{outdomain} 95.43 & \cellcolor{indomain} 12.79 \\
DKT & \cellcolor{outdomain} \textbf{3.49}  & \cellcolor{outdomain} \textbf{4.71} & \cellcolor{outdomain} \textbf{6.61} & \cellcolor{outdomain} \textbf{2.60} & \cellcolor{indomain} \textbf{12.63} \\
\hline
\end{tabular}
\centering
\caption{Comparison with domain generalization methods. The previous methods designed for building domain generalized stereo networks during synthetic data pre-training fail to preserve the domain generalization ability during fine-tuning.}
\label{table: comparison with generalization methods} 
\end{table}

\subsection{Stereo Matching Network Architectures}

In the main paper, we conduct our experiments with robust iterative optimization based stereo matching networks. Here we conduct experiments using other network architectures. We fine-tune CFNet \cite{shen2021cfnet} and CGI-Stereo \cite{xu2023cgi}, which have great domain generalization ability after pre-training on synthetic data. We show the results in \Cref{table: exp on more networks}. Compared to the baseline fine-tuning strategy with GT, networks fine-tuned with DKT show better generalization ability. Furthermore, We explore fine-tuning the recent Croco-Stereo \cite{weinzaepfel2022improved} that builds transformers and train networks with the self-supervised task on a large scale of data. After self-supervised pre-training, Croco-Stereo trains networks to conduct stereo matching jointly on various datasets including SceneFlow, Middlebury, ETH3D, and Booster. We fine-tune Croco-Stereo in the KITTI datasets and evaluate the target and cross domain performance. We note that the cross-domain evaluation in this setting is not to represent the domain generalization ability of the model, but can represent how the model forgets previously seen scenarios. We do not fine-tune Croco-Stereo in the Booster datasets because it has seen the validation set during pre-training.

\begin{table}[h]\footnotesize
\setlength{\abovecaptionskip}{6pt}
\centering
\setlength{\tabcolsep}{4.5pt}
\begin{tabular}{c|rrrrr}
\hline
Method & 2012 & 2015 & Midd & ETH3D & Booster \\
\hline
\hline
Training set & \multicolumn{5}{c}{\textbf{KITTI 2012 \& 2015}} \\
\hline
CFNet \cite{shen2021cfnet} * & \cellcolor{indomain} 5.71 & \cellcolor{indomain} 4.84 & \cellcolor{outdomain} 15.77 & \cellcolor{outdomain} 5.48 & \cellcolor{outdomain} 38.84 \\
CFNet(ft) & \cellcolor{indomain} 2.15 & \cellcolor{indomain} 1.39 & \cellcolor{outdomain} 19.83 & \cellcolor{outdomain} 29.94 & \cellcolor{outdomain} 30.95  \\
DKT-CFNet & \cellcolor{indomain} 2.23 & \cellcolor{indomain} 1.47 & \cellcolor{outdomain} 12.98 & \cellcolor{outdomain} 6.16 & \cellcolor{outdomain} 30.27 \\
\hdashline
CGI-Stereo \cite{xu2023cgi} * & \cellcolor{indomain} 6.55 & \cellcolor{indomain} 5.49  & \cellcolor{outdomain} 13.91 & \cellcolor{outdomain} 6.30 & \cellcolor{outdomain} 33.38 \\
CGI-Stereo(ft) & \cellcolor{indomain} 2.41 & \cellcolor{indomain} 1.58 & \cellcolor{outdomain} 18.62 & \cellcolor{outdomain} 29.84 & \cellcolor{outdomain} 30.51 \\
DKT-CGI & \cellcolor{indomain} 2.26 & \cellcolor{indomain} 1.63 & \cellcolor{outdomain} 14.31 & \cellcolor{outdomain} 7.12 & \cellcolor{outdomain} 29.09 \\
\hdashline
Croco-Stereo \cite{weinzaepfel2022improved} * & \cellcolor{indomain}  12.21 & \cellcolor{indomain}   18.16  &   2.62 &   0.13 &   8.30 \\
Croco-Stereo(ft) & \cellcolor{indomain}  1.81 & \cellcolor{indomain}   1.22  &   7.83  &   2.19 &   23.12 \\
DKT-Croco & \cellcolor{indomain}  1.78 & \cellcolor{indomain}   1.26  &   3.08 &   1.27 &   9.81 \\
\hline
\hline
Training set  & \multicolumn{5}{c}{\textbf{Booster}} \\
\hline
CFNet \cite{shen2021cfnet} * & \cellcolor{outdomain} 4.97 & \cellcolor{outdomain} 6.31 & \cellcolor{outdomain} 15.77 & \cellcolor{outdomain} 5.48 & \cellcolor{indomain} 35.03 \\
CFNet(ft) & \cellcolor{outdomain} 56.20 & \cellcolor{outdomain} 71.41 & \cellcolor{outdomain} 18.45 & \cellcolor{outdomain} 80.53 & \cellcolor{indomain} 25.86  \\
DKT-CFNet & \cellcolor{outdomain} 3.57 & \cellcolor{outdomain} 4.26 & \cellcolor{outdomain} 11.17 & \cellcolor{outdomain} 5.38 & \cellcolor{indomain} 26.11 \\
\hdashline
CGI-Stereo \cite{xu2023cgi} * & \cellcolor{outdomain} 5.90 & \cellcolor{outdomain} 6.02 & \cellcolor{outdomain} 13.91 & \cellcolor{outdomain} 6.30 & \cellcolor{indomain} 30.23 \\
CGI-Stereo(ft) & \cellcolor{outdomain} 30.93 & \cellcolor{outdomain} 46.84 & \cellcolor{outdomain} 20.34 & \cellcolor{outdomain} 46.79 & \cellcolor{indomain} 23.87 \\
DKT-CGI & \cellcolor{outdomain} 5.38 & \cellcolor{outdomain} 5.11 & \cellcolor{outdomain} 13.83 & \cellcolor{outdomain} 6.37 & \cellcolor{indomain} 23.60 \\
\hline
\end{tabular}
\centering
\caption{Results of fine-tuning with more network architectures. * uses pre-trained weights provided by the authors. Our proposed DKT framework can be applied to various network architectures and preserves their domain generalization ability.}
\label{table: exp on more networks} 
\end{table}

\subsection{Fine-tuning on More Datasets}

We perform fine-tuning on the Middlebury and ETH3D datasets, following the data split in MCV-MFC \cite{liang2019stereo}. The experimental results are presented in \Cref{table: exp on more datasets}. Notably, we observe that fine-tuning on these two datasets can lead to a degradation in generalization ability to some unseen domains. However, it's noteworthy that fine-tuning on Middlebury and ETH3D can enhance performance on specific unseen datasets, and overall, the degradation in domain generalization is less pronounced compared to fine-tuning on KITTI and Booster. We think this difference is attributed to the fact that Middlebury and ETH3D datasets contain little transparent or mirrored (ToM) surfaces, which have a substantial impact on degrading domain generalization ability. The modest performance gaps between pre-trained networks and those subjected to fine-tuning suggest that the acquisition of new knowledge during the fine-tuning process may be relatively limited. Moreover, for both datasets, employing DKT during fine-tuning demonstrates better domain generalization ability than using only GT.

\begin{table}[h]\footnotesize
\setlength{\abovecaptionskip}{6pt}
\centering
\begin{tabular}{c|rrrrr}
\hline
Method & 2012 & 2015 & Midd & ETH3D & Booster \\
\hline
\hline
Training set & \multicolumn{5}{c}{\textbf{Middlebury 2014}} \\
\hline
IGEV-Stereo \cite{xu2023iterative} * & \cellcolor{outdomain} 5.13 & \cellcolor{outdomain} 6.04 & \cellcolor{indomain} 5.03 & \cellcolor{outdomain} 3.61 & \cellcolor{outdomain} 17.62 \\
IGEV-Stereo(ft) & \cellcolor{outdomain} 4.02 & \cellcolor{outdomain} 5.01 & \cellcolor{indomain} \textbf{3.81} & \cellcolor{outdomain} 4.97 & \cellcolor{outdomain} 15.26 \\
DKT-IGEV & \cellcolor{outdomain} \textbf{3.47} & \cellcolor{outdomain} \textbf{4.62} & \cellcolor{indomain} 3.83 & \cellcolor{outdomain} \textbf{2.97} & \cellcolor{outdomain} \textbf{14.23} \\
\hline
\hline
Training set  & \multicolumn{5}{c}{\textbf{ETH3D}} \\
\hline
IGEV-Stereo \cite{xu2023iterative} * & \cellcolor{outdomain} 5.13 & \cellcolor{outdomain} 6.04 & \cellcolor{outdomain} \textbf{7.06} & \cellcolor{indomain} 3.09 & \cellcolor{outdomain} 17.62 \\
IGEV-Stereo(ft) & \cellcolor{outdomain} 5.19 & \cellcolor{outdomain} 5.62 & \cellcolor{outdomain} \textcolor{red}{12.31} & \cellcolor{indomain} 2.26 & \cellcolor{outdomain} \textcolor{red}{22.57} \\
DKT-IGEV & \cellcolor{outdomain} \textbf{4.81} & \cellcolor{outdomain} \textbf{5.59} & \cellcolor{outdomain} 7.32 & \cellcolor{indomain} \textbf{2.23} & \cellcolor{outdomain} \textbf{17.33} \\
\hline
\end{tabular}
\centering
\caption{Results of fine-tuning networks on more datasets. * uses pre-trained weights provided by the authors. Networks fine-tuned by the DKT framework show better robustness to unseen domains.}
\label{table: exp on more datasets} 
\end{table}

\begin{table*}[t]\footnotesize
\setlength{\abovecaptionskip}{6pt}
\centering
\setlength{\tabcolsep}{5pt}
\begin{tabular}{c|cccccccccc}
\hline
\multirow{2}{*}{Method} & 2012 & 2015 & Middlebury & ETH3D & Booster & \multicolumn{5}{c}{DrivingStereo}  \\
& \textgreater3px(\%)\ & \textgreater3px(\%)\ & \textgreater2px(\%)\  & \textgreater1px(\%)\ & \textgreater2px(\%)\  & sunny & cloudy & foggy & rainy & avg \\
\hline
CFNet \cite{shen2021cfnet} & \cellcolor{indomain} 2.47 & \cellcolor{indomain} 1.78 & \cellcolor{indomain} 6.96 & \cellcolor{indomain} 1.99 & \cellcolor{indomain} \textcolor{red}{30.43} & \cellcolor{outdomain} 2.75 & \cellcolor{outdomain} 2.49 & \cellcolor{outdomain} 2.03 & \cellcolor{outdomain} \textcolor{red}{6.39} & \cellcolor{outdomain} 3.42 \\
DKT-CFNet & \cellcolor{indomain} 2.51 & \cellcolor{indomain} 1.80 & \cellcolor{indomain} 5.92 & \cellcolor{indomain} \textbf{1.81} & \cellcolor{indomain} 18.55  & \cellcolor{outdomain} 2.20 & \cellcolor{outdomain} 2.34 & \cellcolor{outdomain} 1.89 & \cellcolor{outdomain} 3.55 & \cellcolor{outdomain} 2.50 \\
IGEV-Stereo \cite{xu2023iterative} & \cellcolor{indomain} \textbf{2.00} & \cellcolor{indomain} 1.56 & \cellcolor{indomain} 3.80 & \cellcolor{indomain} 1.98 & \cellcolor{indomain} 12.83 & \cellcolor{outdomain} 2.29 & \cellcolor{outdomain} 1.89 & \cellcolor{outdomain} 1.49 & \cellcolor{outdomain} \textcolor{red}{8.19} & \cellcolor{outdomain} 3.47 \\
DKT-IGEV& \cellcolor{indomain} 2.02 & \cellcolor{indomain} \textbf{1.54} & \cellcolor{indomain} \textbf{3.79} & \cellcolor{indomain} 2.01 & \cellcolor{indomain} \textbf{11.19} & \cellcolor{outdomain} \textbf{2.23} & \cellcolor{outdomain} \textbf{1.81} & \cellcolor{outdomain} \textbf{1.42} & \cellcolor{outdomain} \textbf{3.31} & \cellcolor{outdomain} \textbf{2.19}  \\
\hline
\end{tabular}
\centering
\caption{Results of joint generalization. Networks are fine-tuned on a combination of KITTI 2012, KITTI 2015, Middlebury, ETH3D, and Booster datasets. Networks fine-tuned by the DKT framework show competitive joint generalization performance, as well as better robustness to unseen challenging weather.}
\label{supp_table: joint generalization} 
\end{table*}

\subsection{Joint Generalization}

In addition to fine-tuning stereo matching networks on individual real-world scenarios, we employ DKT for joint fine-tuning across multiple domains. Besides assessing performance in target domains, we also evaluate the domain generalization ability on previously unseen DricingStereo scenarios. The results, presented in \Cref{supp_table: joint generalization}, demonstrate that using DKT for joint fine-tuning yields comparable results across multiple seen domains, while exhibiting superior robustness on unseen scenarios.

\section{Additional Qualitative Results} \label{supp_sec: more visualization results}

\begin{figure*}[h]
\centering
\includegraphics[width=1.0\linewidth]{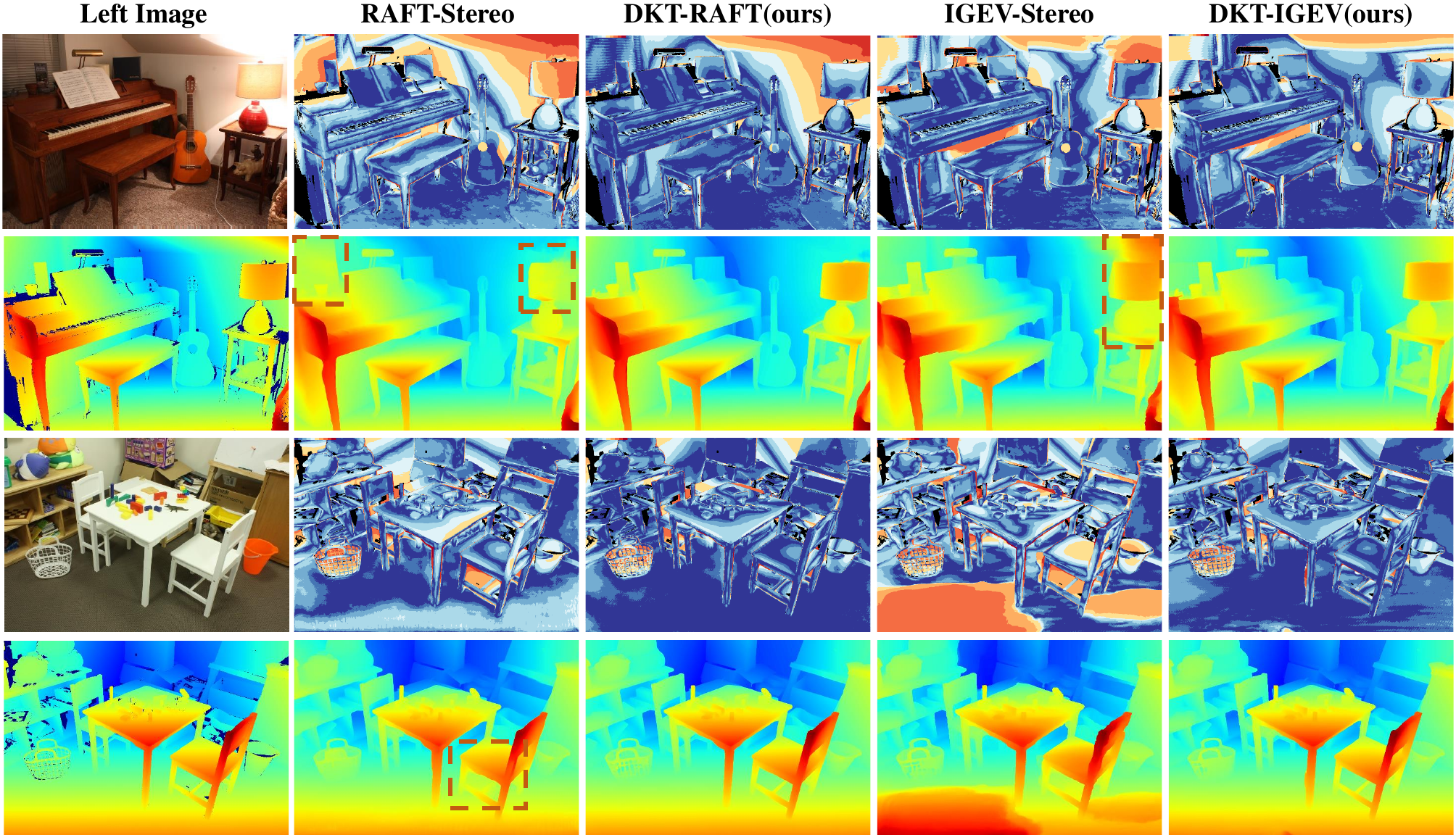}
\vspace{-6mm}
\caption{Qualitative results of KITTI fine-tuned networks on the Middlebury training set. The left panel shows the left input image and the ground truth disparity. For each example, the first row shows the error map and the second row shows the colorized disparity prediction.}
\label{supp_fig: kittift_midd}
\end{figure*}

\begin{figure*}[h]
\centering
\includegraphics[width=1.0\linewidth]{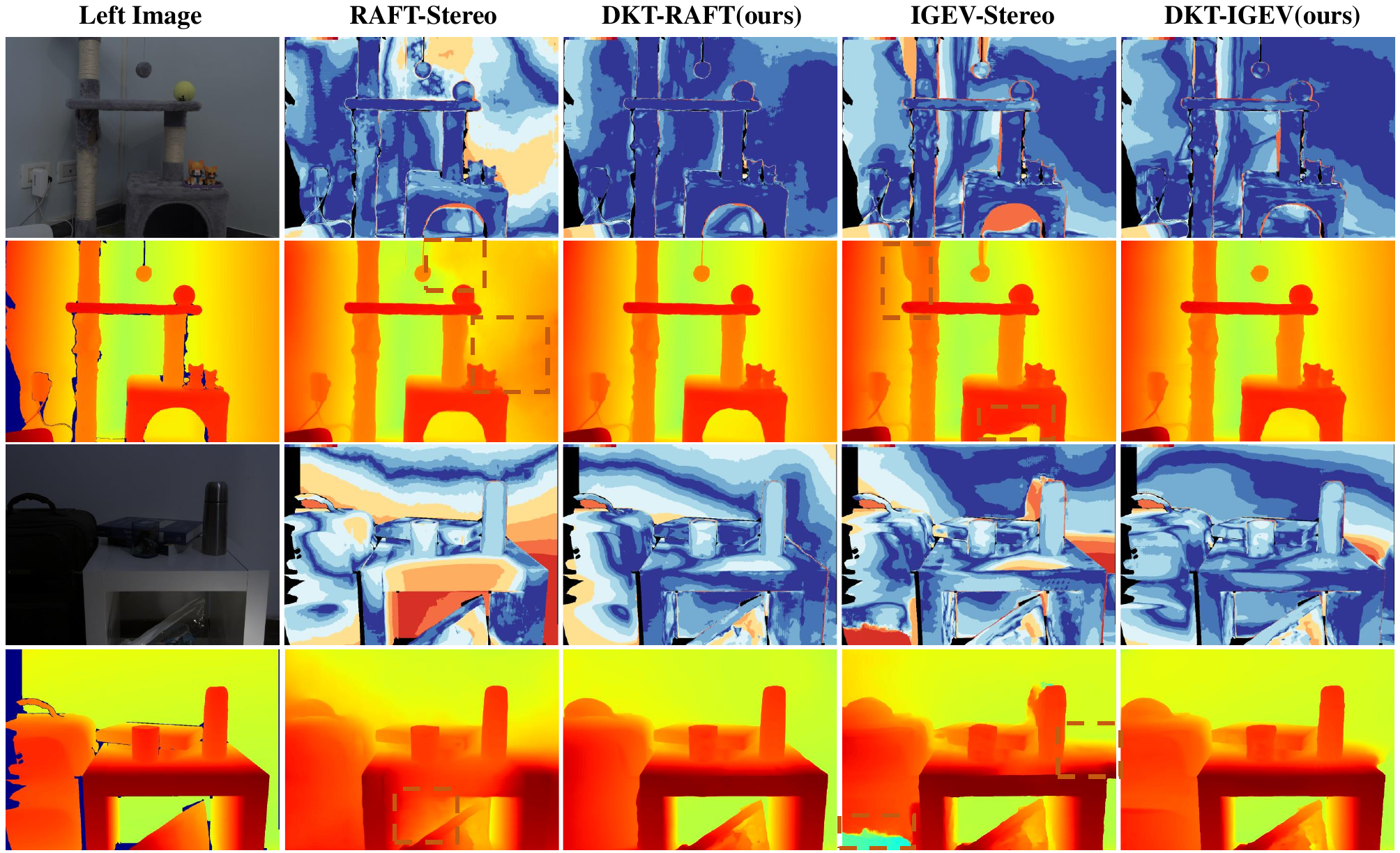}
\caption{Qualitative results of KITTI fine-tuned networks on the Booster training set. The left panel shows the left input image and the ground truth disparity. For each example, the first row shows the error map and the second row shows the colorized disparity prediction.}
\vspace{-6mm}
\label{supp_fig: kittift_booster}
\end{figure*}

\begin{figure*}[h]
\centering
\includegraphics[width=1.0\linewidth]{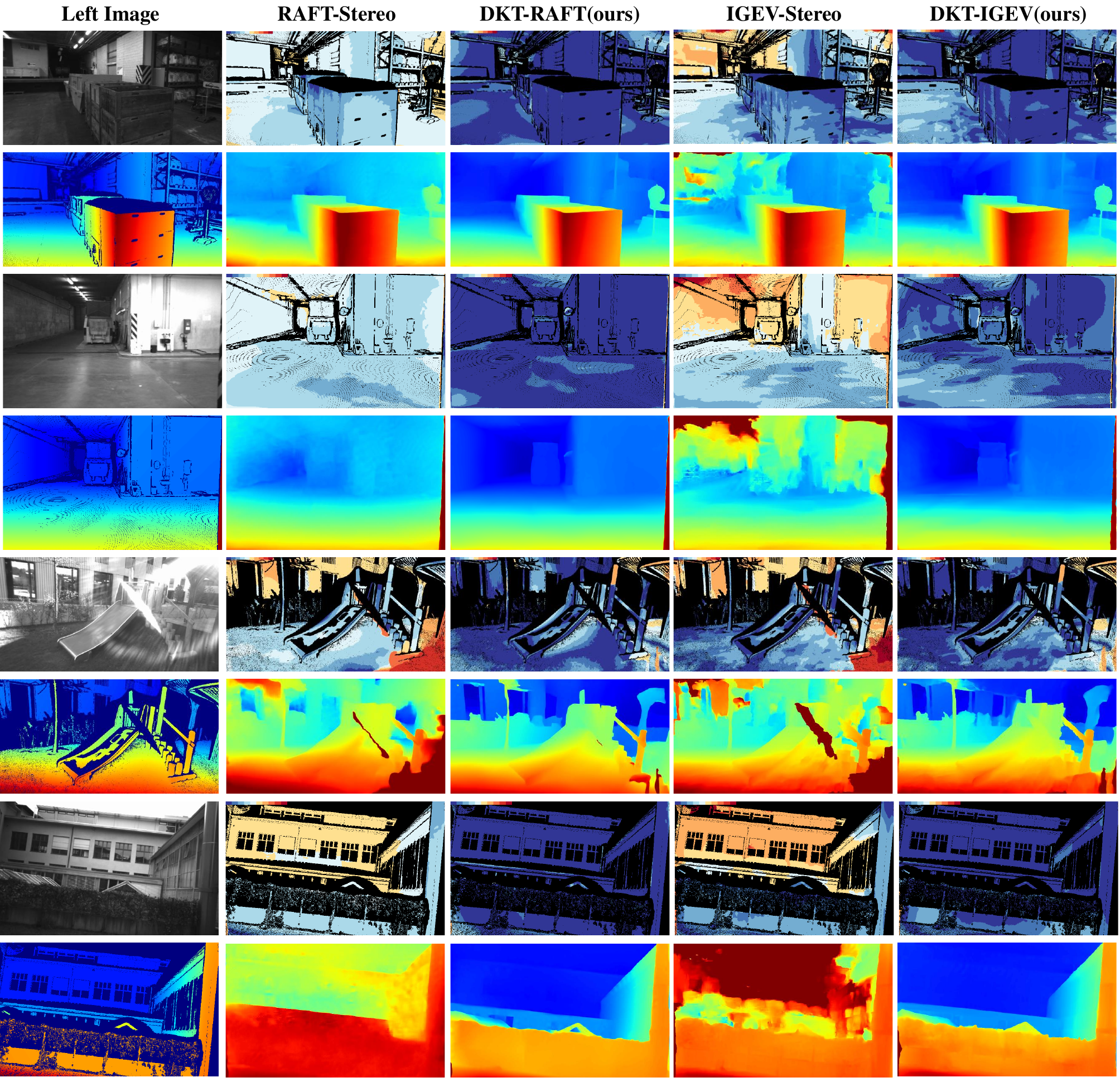}
\caption{Qualitative results of KITTI fine-tuned networks on the ETH3D training set. The left panel shows the left input image and the ground truth disparity. For each example, the first row shows the error map and the second row shows the colorized disparity prediction.}
\label{supp_fig: kittift_eth3d}
\end{figure*}

\begin{figure*}[h]
\centering
\includegraphics[width=1.0\linewidth]{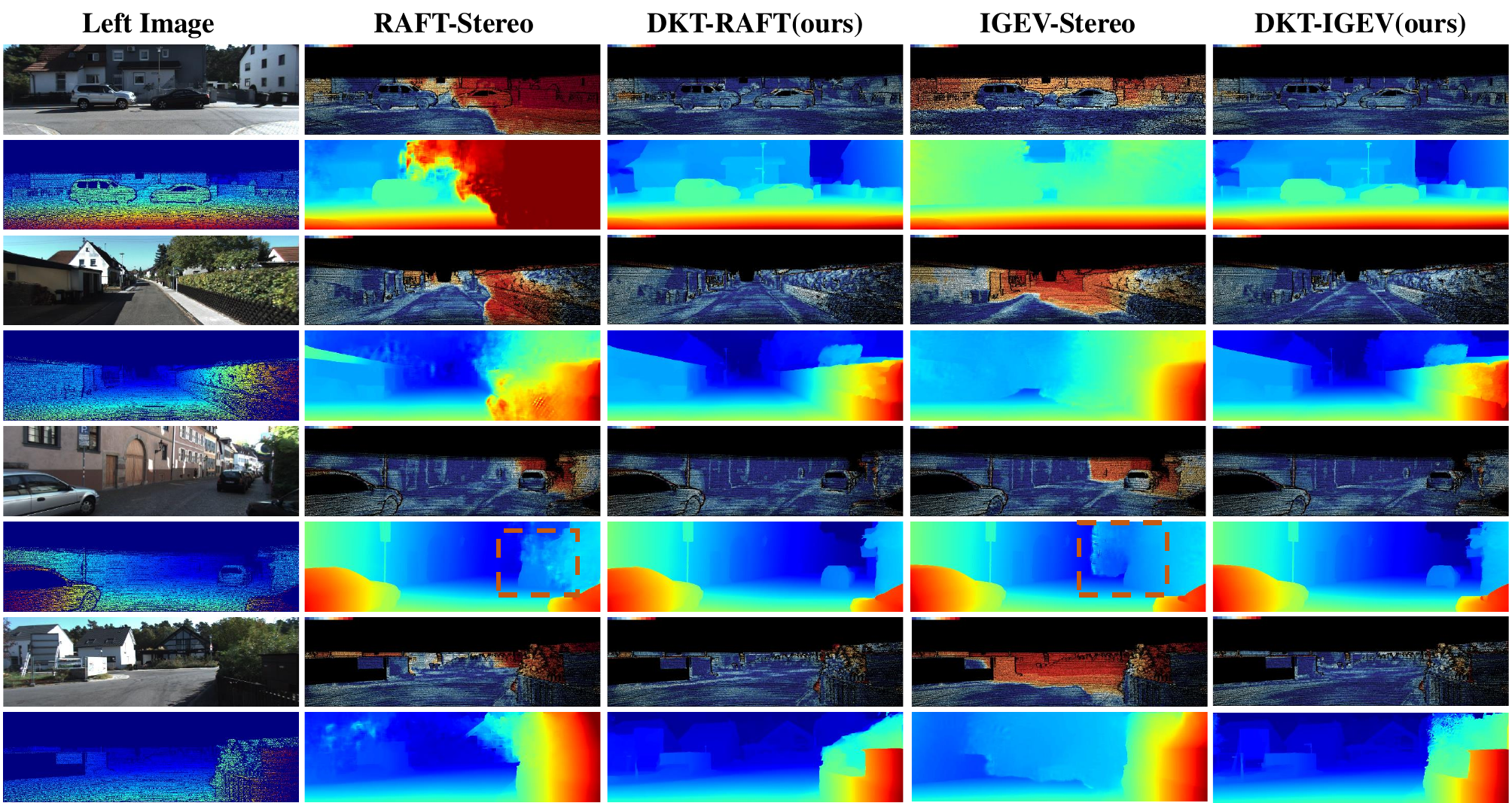}
\caption{Qualitative results of Booster fine-tuned networks on the KITTI 2012 training set. The left panel shows the left input image and the ground truth disparity. For each example, the first row shows the error map and the second row shows the colorized disparity prediction.}
\label{supp_fig: boosterft_kt2012}
\end{figure*}

\begin{figure*}[h]
\centering
\includegraphics[width=1.0\linewidth]{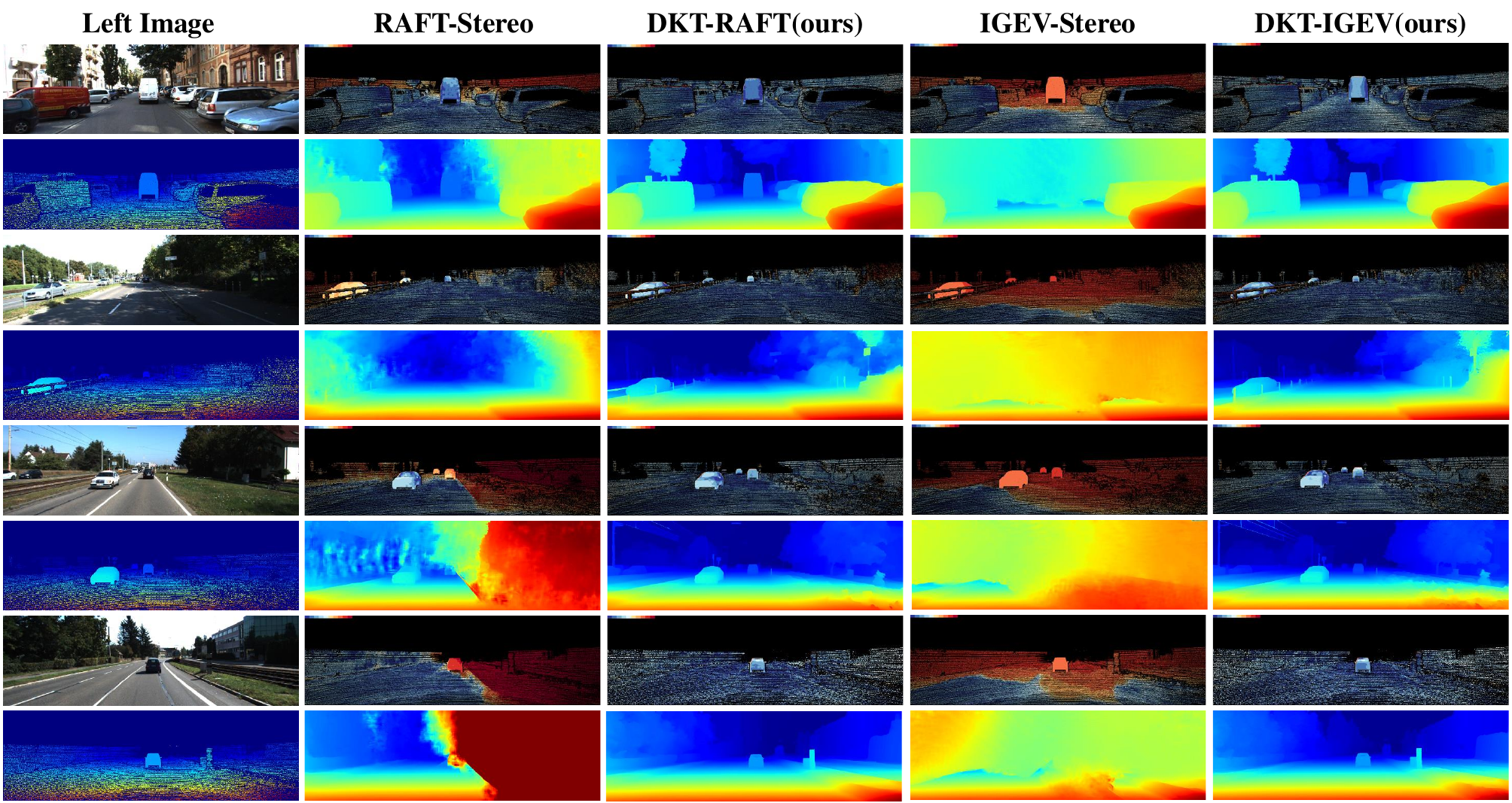}
\caption{Qualitative results of Booster fine-tuned networks on the KITTI 2015 training set. The left panel shows the left input image and the ground truth disparity. For each example, the first row shows the error map and the second row shows the colorized disparity prediction.}
\label{supp_fig: boosterft_kt2015}
\end{figure*}

\begin{figure*}[h]
\centering
\includegraphics[width=1.0\linewidth]{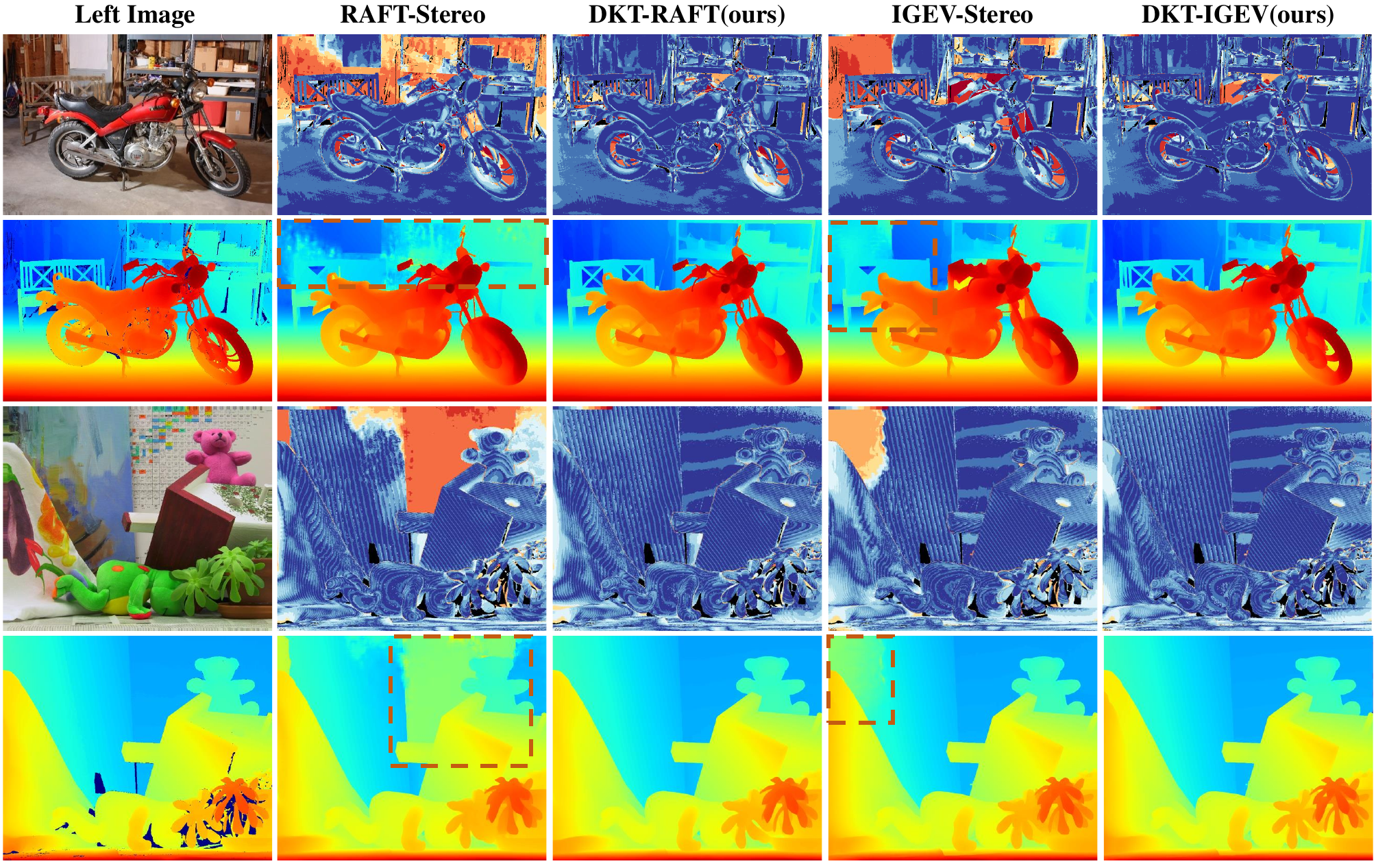}
\caption{Qualitative results of Booster fine-tuned networks on the Middlebury training set. The left panel shows the left input image and the ground truth disparity. For each example, the first row shows the error map and the second row shows the colorized disparity prediction.}
\label{supp_fig: boosterft_midd}
\end{figure*}

\begin{figure*}[h]
\centering
\includegraphics[width=1.0\linewidth]{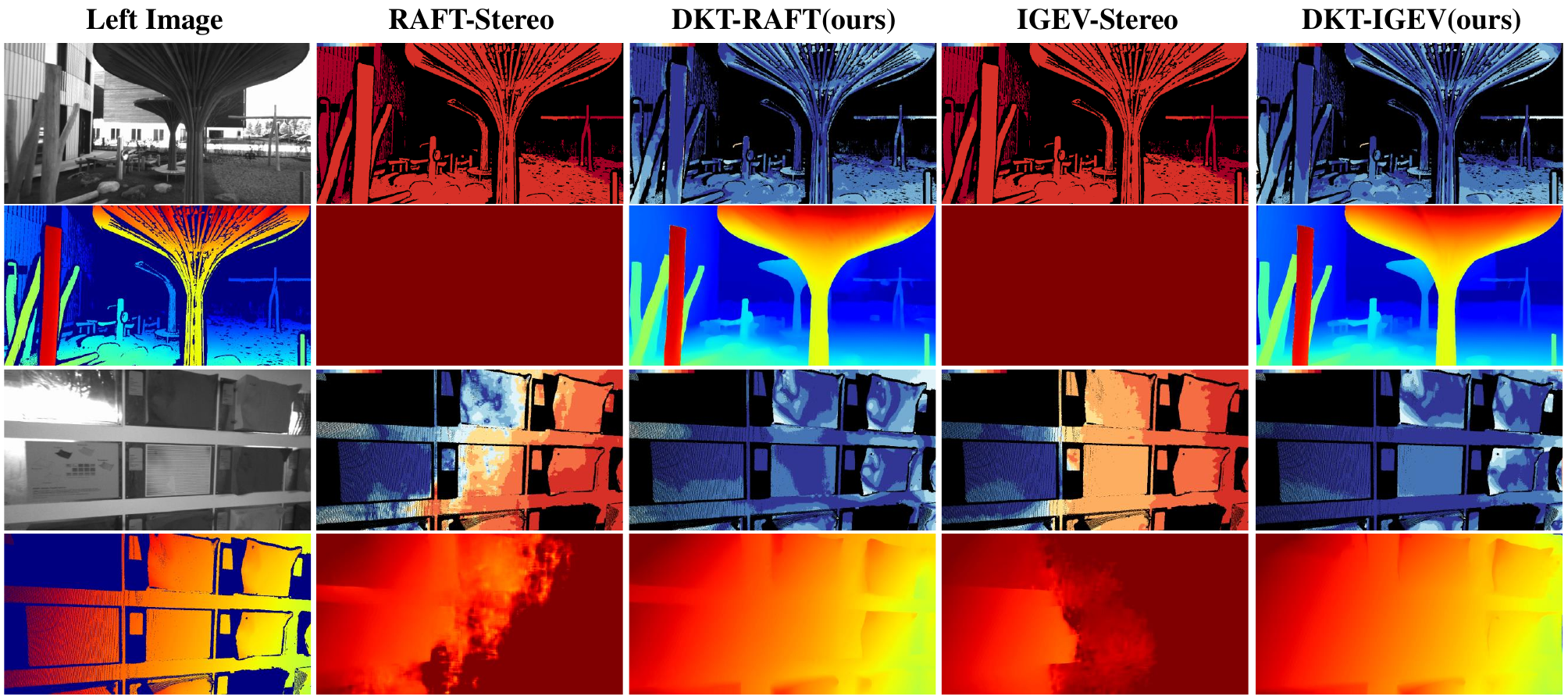}
\caption{Qualitative results of Booster fine-tuned networks on the ETH3D training set. The left panel shows the left input image and the ground truth disparity. For each example, the first row shows the error map and the second row shows the colorized disparity prediction.}
\label{supp_fig: boosterft_eth3d}
\end{figure*}

In this section, we provide additional qualitative results of the domain generalization performance of stereo matching networks fine-tuned with only GT and DKT. Compared to using only GT for fine-tuning, DKT effectively preserves the networks' robustness to unseen domains after fine-tuning. \Cref{supp_fig: kittift_midd,supp_fig: kittift_booster,supp_fig: kittift_eth3d} use the same networks fine-tuned on the KITTI datasets and show the performance on unseen Middlebury, Booster, and ETH3D domains. \Cref{supp_fig: boosterft_kt2012,supp_fig: boosterft_kt2015,supp_fig: boosterft_midd,supp_fig: boosterft_eth3d} use the same networks fine-tuned on the Booster dataset and show the performance on unseen KITTI 2012, KITTI 2015, Middlebury, and ETH3D domains.

\clearpage

\clearpage

\clearpage
{
    \small
    \bibliographystyle{ieeenat_fullname}
    \bibliography{main}
}

\end{document}